\documentclass[journal]{IEEEtran}
\IEEEoverridecommandlockouts
\usepackage{cite}
\usepackage{multirow} 
\usepackage{amsmath,amssymb,amsfonts,mathtools,bm}
\usepackage{amsthm}
\usepackage{algorithm}
\usepackage{algorithmic}
\usepackage{graphicx}
\usepackage{textcomp}
\usepackage{xcolor,float}
\usepackage{tabularx}
\usepackage{textcomp}
\usepackage{diagbox}
\usepackage{svg}
\usepackage{booktabs}
\usepackage{hyperref}
\usepackage{pifont}

\usepackage{threeparttable}
\usepackage{booktabs}
\usepackage{tabularx}
\usepackage{subfig}
\usepackage[utf8]{inputenc}
\usepackage[T1]{fontenc}
\usepackage{float}
\usepackage[labelformat=empty]{subcaption} 
\usepackage{graphicx}
\usepackage{multirow} 
\usepackage{siunitx}
\usepackage{subfig}
\usepackage{placeins}
\usepackage{orcidlink}
\usepackage[final]{changes} 

\allowdisplaybreaks
\renewcommand{\baselinestretch}{1}

\usepackage{orcidlink}  

\newcommand{\orcidauthorA}{0009-0004-4048-4668}
\newcommand{\orcidauthorB}{0000-0003-1439-4875}
\newcommand{\orcidauthorC}{0000-0001-7907-2071}
\newcommand{\orcidauthorD}{0000-0003-3056-9030}
\newcommand{\orcidauthorE}{0000-0002-2062-8327}
\newcommand{\orcidauthorF}{0000-0002-5207-6504}
\newcommand{\orcidauthorG}{0000-0003-2608-775X}
\newcommand{\orcidauthorH}{0000-0002-4140-287X}

\begin{document}
\title{RadioDiff-3D: A 3D$\times$3D Radio Map Dataset and Generative Diffusion Based Benchmark for 6G Environment-Aware Communication}


\author{
Xiucheng Wang~\orcidlink{\orcidauthorB},~\IEEEmembership{Graduate Student Member,~IEEE,}
Qiming Zhang~\orcidlink{\orcidauthorA},
Nan Cheng~\orcidlink{\orcidauthorC},~\IEEEmembership{Senior Member,~IEEE,}
Junting Chen~\orcidlink{\orcidauthorD},~\IEEEmembership{Member,~IEEE,}
Zezhong Zhang~\orcidlink{\orcidauthorE},~\IEEEmembership{Member,~IEEE,}
Zan Li~\orcidlink{\orcidauthorF},~\IEEEmembership{Fellow,~IEEE,}\\
Shuguang Cui~\orcidlink{\orcidauthorG},~\IEEEmembership{Fellow,~IEEE,}
and Xuemin (Sherman) Shen~\orcidlink{\orcidauthorH},~\IEEEmembership{Fellow,~IEEE}

\thanks{
\par This work was supported by the National Key Research and Development Program of China (2024YFB907500).
\par Xiucheng Wang, Nan Cheng, and Zan Li are with the State Key Laboratory of ISN and School of Telecommunications Engineering, Xidian University, Xi’an 710071, China (e-mail: xcwang\_1@stu.xidian.edu.cn; dr.nan.cheng@ieee.org; zanli@xidian.edu.cn);\textit{(Xiucheng Wang and Qiming Zhang contributed equally to this work.)(Corresponding author: Nan Cheng.)}. 
\par Qiming Zhang is with the School of Artificial Intelligence, Xidian University, Xi’an 710071, China (e-mail: 23009200991@stu.xidian.edu.cn).
\par Junting Chen, Zezhong Zhang and Shuguang Cui are with the School of Science and Engineering (SSE), Shenzhen Future Network of Intelligence Institute (FNii-Shenzhen), and Guangdong Provincial Key Laboratory of Future Networks of Intelligence, The Chinese University of Hong Kong (Shenzhen), Shenzhen, China (e-mail: shuguangcui@cuhk.edu.cn; juntingc@cuhk.edu.cn; zhangzezhong@cuhk.edu.cn);
\par Xuemin (Sherman) Shen is with the Department of Electrical and Computer Engineering, University of Waterloo, Waterloo, N2L 3G1, Canada (e-mail: sshen@uwaterloo.ca).
}     
} 
    
    \maketitle

\IEEEdisplaynontitleabstractindextext

\IEEEpeerreviewmaketitle

\begin{abstract}
Radio maps (RMs) serve as a critical foundation for enabling environment-aware wireless communication, as they provide the spatial distribution of wireless channel characteristics. Despite recent progress in RM construction using data-driven approaches, most existing methods focus solely on pathloss prediction in a fixed 2D plane, neglecting key parameters such as direction of arrival (DoA), time of arrival (ToA), and vertical spatial variations. Such a limitation is primarily due to the reliance on static learning paradigms, which hinder generalization beyond the training data distribution. To address these challenges, we propose UrbanRadio3D, a large-scale, high-resolution 3D RM dataset constructed via ray tracing in realistic urban environments. \added{UrbanRadio3D is over 37× larger than previous datasets across a 3D space with 3 metrics as pathloss, DoA, and ToA, forming a novel 3D×3D dataset with 7× more height layers than prior state-of-the-art (SOTA) dataset}. To benchmark 3D RM construction, a UNet with 3D convolutional operators is proposed. Moreover, we further introduce RadioDiff-3D, a diffusion-model-based generative framework utilizing the 3D convolutional architecture. RadioDiff-3D supports both radiation-aware scenarios with known transmitter locations and radiation-unaware settings based on sparse spatial observations. Extensive evaluations on UrbanRadio3D validate that RadioDiff-3D achieves superior performance in constructing rich, high-dimensional radio maps under diverse environmental dynamics. This work provides a foundational dataset and benchmark for future research in 3D environment-aware communication. The dataset is available at \url{https://github.com/UNIC-Lab/UrbanRadio3D}.
\end{abstract}


\section{Introduction}
To support the increasing demands of immersive communications and ultra-reliable low-latency services in sixth-generation (6G) networks, it is imperative to significantly enhance spectrum efficiency and network coverage \cite{6g,shen2023toward,9511234-tnse-6g,9674855-tnse-6g}. This objective is driving the evolution of network architecture toward extremely large multiple-input multiple-output (XL-MIMO) systems, which feature antenna arrays with over 1024 elements \cite{han2020channel,9851495-tnse-mimo,luo2018channel-mimo}, and ultra-dense networks (UDNs) with densely deployed access points \cite{Liu2023d}. However, this evolution imposes unprecedented challenges for traditional pilot-based channel estimation, as the volume of channel state information (CSI) to be acquired grows prohibitively large \cite{han2020channel}. In XL-MIMO scenarios, pilot transmission and estimation may consume over 90\% of the total time slot, severely undermining the goal of frequency efficiency and rendering real-time adaptation infeasible \cite{wang2024tutorial}. Concurrently, to expand coverage and support dynamic user distributions, 6G networks will incorporate numerous mobile access nodes, such as autonomous aerial vehicles (AAVs) and low-Earth-orbit satellites \cite{Liu2023d,9159929-tnse-uav,9215008-tnse-uav,9162508-tnse-uav}. The mobility of these nodes demands proactive trajectory planning to maintain reliable connectivity and quality-of-service (QoS). This, in turn, requires advanced knowledge of spatial channel variations prior to node arrival \cite{cheng2019space}. Collectively, these trends call for a paradigm shift from pilot-reliant mechanisms toward environment-aware communication, where ambient information is leveraged to infer channel characteristics without relying on direct measurements. In this context, the radio map (RM) has emerged as a key enabler, providing a precomputed spatial representation of wireless channel features, such as pathloss, that can guide real-time network optimization and trajectory control in complex 6G environments \cite{zeng2021toward,zeng2024tutorial}.

\begin{table*}[t]
\centering
\caption{Comparison of Different RM Datasets.}
\vspace{-2pt}
\resizebox{0.9\linewidth}{!}{
\begin{tabular}{@{}c|c|c|c|c|c@{}}
\toprule
Properties  & UrbanRadio3D (Ours) & SpetrumNet\cite{zhang2024generative} & RadioMapSeer \cite{levie2021radiounet} & RadioGAT \cite{li2024radiogat} & CKMImageNet \cite{wu2024ckmimagenet}   \\ \hline 
Dataset Size & {\color[HTML]{9A0000} \textbf{11.2M}}   & {\color[HTML]{00009B} \underline{300k}}  & 56k & 21K  & 72k  \\ \hline 
Number of Height & {\color[HTML]{9A0000} \textbf{20}} & {\color[HTML]{00009B} \underline{3}}  & 1 & 1   & 1  \\ \hline 
Number of Map & {\color[HTML]{00009B} \underline{701}}  & {\color[HTML]{9A0000} \textbf{764}}  & 701 & 10 & 42 \\ \hline 
Number of BS Location & {\color[HTML]{9A0000} \textbf{200}} & 4  & {\color[HTML]{00009B} \underline{80}}  & 3 & 1$\sim$42  \\ \hline 
Size of Map  & {\color[HTML]{9A0000} \textbf{256$\times$256}} & 128$\times$128  & {\color[HTML]{00009B} \underline{256$\times$256}} & 200$\times$200  & 128$\times$128  \\  \hline 
Horizontal Resolution & {\color[HTML]{9A0000} \textbf{1 m}}     & 10 m       & {\color[HTML]{00009B} \underline{1 m}}          & 5 m      & 2 m         \\ \hline 
Height Resolution & {\color[HTML]{9A0000} \textbf{1 m}}   & N/A  & N/A  & N/A  & N/A   \\  \hline 
Real Buildings Shape  & {\color[HTML]{9A0000} \ding{52}}   & \ding{52}   & \ding{52}  & \ding{52}   & \ding{52} \\  \hline 
Real Buildings Height & {\color[HTML]{9A0000} \ding{52}}   & {\color[HTML]{00009B} \underline{\ding{52}}}   & \ding{55}   & \ding{55}   & \ding{52}  \\ \hline 
Pathloss & {\color[HTML]{9A0000} \ding{52}}    & \ding{52}   & \ding{52}  & \ding{52}   & \ding{52}   \\  \hline 
DoA\_Azi  & {\color[HTML]{9A0000} \ding{52}}   & \ding{55}   & \ding{55}  & \ding{55}  & {\color[HTML]{00009B} \underline{Few}}   \\  \hline 
DoA\_Ele   & {\color[HTML]{9A0000} \ding{52}}  & \ding{55}  & \ding{55}    & \ding{55}    & {\color[HTML]{00009B} \underline{Few}}  \\ 
\bottomrule
\end{tabular}
}
\label{tab1}
\vspace{-12pt}
\end{table*}

Despite the benefits of radio maps for enabling environment-aware communications, their construction remains a technically demanding task \cite{zeng2024tutorial}. High-fidelity RM generation traditionally relies on numerically solving Maxwell’s equations or the Helmholtz wave equation using techniques such as finite-difference time-domain (FDTD) methods \cite{jones2013theory}. However, the computational complexity of FDTD is prohibitively high, restricting its practical application to regions spanning only a few electromagnetic wavelengths\cite{jones2013theory}. Even with approximate techniques like electromagnetic ray tracing (ERT), the generation of RMs over street-scale urban environments often requires tens of minutes per instance \cite{deschamps1972ray}, rendering these methods unsuitable for real-time or near-real-time inference in dynamic 6G scenarios. In response, neural network (NN)-based approaches have garnered significant attention due to their potential to infer RMs with high speed and reasonable accuracy \cite{levie2021radiounet}. However, current NN-based methods—including convolutional neural networks (CNNs) \cite{levie2021radiounet,zhang2023rme}, graph neural networks (GNNs) \cite{chen2023graph}, and even state-of-the-art generative large model based methods \cite{wang2024radiodiff}, are predominantly constrained to constructing 2D pathloss distributions at a fixed height, which is typically 1.5 meters. This limitation stems from the lack of publicly available datasets containing diverse and densely sampled 3D radio environment data. According to the principles of statistical learning theory, neural networks are inherently constrained to the data distribution present in their training sets, and thus fail to generalize across unrepresented vertical or directional dimensions \cite{vapnik1998statistical}. Consequently, existing models offer limited utility in applications that demand volumetric RF knowledge, such as precise beamforming, user positioning, or real-time 3D coverage planning. More critically, their inability to generate 3D radio maps severely limits their applicability in safety-critical tasks such as AAV navigation \cite{qiu2024channel,wang2022joint}. In practice, AAVs often operate in highly dynamic airspace over unlicensed industrial scientific medical (ISM) bands, which are prone to strong and heterogeneous interference \cite{qiu2024channel}. Therefore, without access to spatially rich 3D RM data, trajectory planning becomes suboptimal or even hazardous. Thus, there is a compelling need for a new generation of RM datasets and models that move beyond 2D pathloss predictions—enabling robust, multi-dimensional inference across 3D space to support intelligent decision-making in advanced 6G applications.
\begin{table*}[t]
\centering
\caption{Comparison of Different RM Construction Methods.}
\vspace{-2pt}
\resizebox{0.9\linewidth}{!}{
\begin{tabular}{@{}c|c|c|c|c|c@{}}
\toprule
Properties  & RadioDiff-3D (Ours) & RadioDiff\cite{wang2024radiodiff} & RadioUNet \cite{levie2021radiounet} & RME-GAN \cite{zhang2023rme} & LocUNet \cite{yapar2022locunet}   \\ \hline 
PathLoss Prediction & {\color[HTML]{9A0000} \ding{52}}   & \ding{52}  & \ding{52} & \ding{52}  & \ding{52}  \\ \hline 
ToA Prediction  & {\color[HTML]{9A0000} \ding{52}} & \ding{55}  & \ding{55} & \ding{55}   & \ding{52}  \\ \hline 
DoA\_Ele Prediction & {\color[HTML]{9A0000} \ding{52}} & \ding{55}  & \ding{55} & \ding{55}   & \ding{55}  \\ \hline 
DoA\_Azi Prediction & {\color[HTML]{9A0000} \ding{52}} & \ding{55}  & \ding{55} & \ding{55}   & \ding{55}  \\ \hline 
Sampling Information & Alternative     & Cannot       & Alternative  & Necessary     & Cannot \\ \hline 
Environment Dimmention & {\color[HTML]{9A0000} \textbf{3D}} & 2D  & 2D  & 2D & 2D   \\  \hline 
Generative AI  & {\color[HTML]{9A0000} \ding{52}} & \ding{52}  & \ding{55} & \ding{55}   & \ding{55} \\  
\bottomrule
\end{tabular}
}
\label{tab2}
\vspace{-12pt}
\end{table*}

To address the aforementioned limitations, this paper introduces a comprehensive large-scale 3D×3D RM dataset, constructed using a ray-tracing-based electromagnetic simulation pipeline grounded in the realistic height and geometry of urban buildings. With a fine-grained spatial resolution of one cubic meter, this dataset comprises over ten million labeled data points, exceeding the scale of existing RM datasets by more than \textbf{37×}. Different from prior datasets that are restricted to 2D pathloss information at fixed heights \cite{levie2021radiounet,li2024radiogat}, our proposed dataset captures rich multi-dimensional channel characteristics, including pathloss, direction of arrival (DoA) in both azimuth and elevation, and time of arrival (ToA). This detailed representation of spatial propagation forms a foundational benchmark for advancing 3D-aware wireless intelligence. Furthermore, the volumetric nature of the data reveals complex spatial correlations not only across adjacent grid points on a plane but also along the vertical dimension, enabling learning models to exploit inter-layer dependencies for more accurate and efficient RM construction. To fully leverage these properties, we propose a novel generative framework, RadioDiff-3D, which adopts a denoising diffusion probabilistic model embedded with 3D convolutional architectures. Unlike traditional neural networks that infer RMs slice-by-slice in 2D, RadioDiff-3D operates directly in 3D space to synthesize high-fidelity RMs across varying heights. Moreover, the model is designed to handle both radiation-aware and radiation-unaware scenarios: in the former, it constructs 3D RMs for cooperative base stations using known environmental and transmitter information; in the latter, it estimates interference distributions from non-cooperative transmitters using sparsely sampled observations and environmental priors. This dual-mode generative capability not only expands the applicability of RM construction to more realistic and complex scenarios, but also provides essential situational awareness for critical applications such as AAV trajectory optimization and interference-avoidance navigation in contested spectral environments. The main contributions of this paper are summarized as follows.
\begin{enumerate}
    \item Different from prior RM datasets such as RadioMapSeer that focus primarily on 2D pathloss distributions at fixed heights, \added{UrbanRadio3D introduces a 3D×3D channel representation that characterizes the spatial distributions of 3 metrics as pathloss, DoA, and ToA across a 3D environmental distribution}. This multi-modal and spatially continuous structure enables in-depth modeling of elevation-sensitive propagation effects, supporting advanced tasks such as 3D localization, altitude-aware beamforming, and volumetric coverage optimization in next-generation wireless networks.
    \item UrbanRadio3D introduces a large-scale, spatially resolved dataset with 1-meter cubic resolution, generated using high-fidelity ray tracing over realistic urban geometries with diverse building heights and layouts. The dataset is organized in a consistent voxel-based format across multiple modalities and receiver altitudes, bridging structured electromagnetic simulation data with 3D convolutional and diffusion-based AI models. This design facilitates efficient integration into learning-based volumetric reconstruction frameworks and promotes reproducibility across wireless AI research.
    \item Two benchmark models for 3D radio map (RM) construction have been proposed in this paper. The first is a convolutional baseline that employs a UNet architecture with 3D convolutional operators, serving as a representative benchmark for volumetric CNN-based methods. The second is a generative diffusion model, termed RadioDiff-3D, which leverages 3D convolutions to synthesize high-fidelity radio maps across the full spatial volume. By modeling the joint spatial dependencies throughout three dimensions, RadioDiff-3D enables the reconstruction of dense, multi-features 3D channel representations and extends beyond the limitations of fixed-height prediction commonly found in prior works.
\end{enumerate}

\section{Related Works and Preliminary}
\subsection{RM Construction Methods}
RM construction has progressed from early measurement-driven interpolation to the latest fully generative, environment-aware synthesis. For clarity, we group prior work into two trajectories: sampling-based inference, which reconstructs the field from sparse path-loss measurements (SPM), and sampling-free inference, which exploits environmental priors and data-driven models to bypass on-site probing. A detailed review of both strands is essential for appreciating the technical positioning of our contribution.

The archetypal workflow begins with a calibrated scanner traversing the region of interest to collect SPM, followed by spatial interpolation. Early cellular deployments relied on inverse-distance weighting (IDW) and K-nearest-neighbour (KNN) interpolation, where the estimate at an unmeasured point is a convex combination of its K closest observations whose weights decay with Euclidean distance \cite{144,145}. Although computationally trivial, these schemes treat path loss as an isotropic field and neglect the anisotropy introduced by streets, building façades, and foliage, leading to appreciable error once the sampling density falls below roughly one point per lamppost in dense urban trials \cite{148}. To encode local curvature, researchers adopted local polynomial regression—also called local multinomial regression, where a first or second-order surface is fit to neighbourhood samples by weighted least squares \cite{146,147}. This reduces bias on gentle gradients but still ignores cross-neighbourhood correlation. A more rigorous statistical foundation is offered by Kriging: ordinary Kriging assumes second-order stationarity, estimates the semivariogram from data, and derives the best linear unbiased predictor that minimises the mean-squared error (MSE) under that covariance model \cite{141,142}. Kriging is optimal within its linear subspace, yet its matrix inversion costs $\mathcal{O}(N^3)$ and becomes burdensome when thousands of samples are available. To reconcile sparsity with long-range correlation, matrix and tensor completion discretize the region into a grid and exploit the empirical low rank of large-scale path-loss matrices \cite{159,160}. Nuclear-norm minimisation successfully recovers entire city-blocks from <10\% samples when the observation pattern satisfies incoherence conditions. However, identifiability breaks down for irregular sampling; interpolation-assisted completion augments missing pixels with local Kriging estimates to maintain recoverability while preserving global low rank \cite{159}. Kernel regression in reproducing-kernel Hilbert space connects Kriging to Gaussian-process regression and offers closed-form solutions with dual-kernel designs that separate slowly varying path-loss trends from fast local shadowing \cite{157}. Temporal dynamics introduce additional complications. Kriged Kalman filtering fuses time-varying measurements with spatial Kriging under a state–space formulation, yielding recursive predictors that track moving obstacles and seasonal changes \cite{141}. While effective, the filter inherits Kriging’s cubic complexity in the state dimension. Despite decades of refinement, sampling-based inference retains two structural weaknesses. First, its dependence on SPM makes it unsuitable for safety-critical facilities, disaster zones, or aerial corridors where drive-testing is impractical. Second, interpolation error scales super-linearly with inter-sample spacing in NLOS regions, causing rapid degradation once sampling density drops below a terrain-specific threshold—an effect verified in the “Interference Cartography Manager” field campaign \cite{148,149}.

To circumvent these limitations, sampling-free methods condition on a priori information such as digital surface models (DSM), building footprints, BS metadata, and land-use labels. The earliest embodiment is deterministic ray tracing, which shoots electromagnetic rays through a 3-D model to compute the path-loss map. Although high fidelity is achievable with modern multi-bounce solvers, both data acquisition and computation remain prohibitive at city scale \cite{137,138}. Consequently, attention shifted to statistical surrogates that learn an RM generator directly from coarse environmental rasters. A landmark in this space is RadioUNet, which feeds a binary raster of building masks into a U-Net and regresses the 2-D path-loss field using pixelwise MSE loss \cite{levie2021radiounet}. RadioUNet significantly outperforms IDW and Kriging in unseen European downtowns, but its receptive field, while multiscale, is still ultimately local. RadioNet extends RadioUNet by inserting transformer attention layers after each U-Net bottleneck, allowing the network to learn long-range diffraction correlations—e.g., shadowing that persists across parallel streets separated by courtyards—thereby boosting accuracy on kilometre-scale maps \cite{150}. Recognising that city topology is naturally graph-structured, graph neural networks (GNNs) have been proposed for RM prediction. In GAT-REM and GraphREM, buildings, streets, and BSs become nodes, edges encode line-of-sight or first-order diffraction relationships, and attention-weighted message passing propagates features across the graph, yielding robust generalisation to new cities with unseen block layouts \cite{166,167}. Nevertheless, these models remain discriminative: they output a single deterministic map for a given environment and cannot express the inherent uncertainty of propagation in dynamic or partially known scenes. Generative paradigms attempt to close this expressiveness gap. RME-GAN augments a convolutional generator with an adversarial discriminator so that the synthesised RM resembles the empirical distribution of true maps while respecting an MSE reconstruction term \cite{zhang2023rme}. Although adversarial supervision sharpens spatial textures, RME-GAN still requires sparse SPM as conditional anchors, precluding true sampling-free deployment. A decisive break with measurement dependency is offered by the RadioDiff family. RadioDiff \cite{wang2024radiodiff} trains a denoising diffusion probabilistic model (DDPM) to reverse a Markovian Gaussian perturbation process conditioned solely on building rasters and BS coordinates, enabling RM synthesis directly from environmental data. RadioDiff-$k^2$ \cite{wang2025radiodiffk} introduces a score-based “knowledge transfer” step that couples synthetic ray-traced data with limited real measurements, thereby bridging simulation-to-reality gaps, while RadioDiff-Inv \cite{wang2025radiodiffinverse} leverages the same diffusion kernel for inverse problems, such as inferring missing BS positions from partial RMs. These diffusion models achieve new state-of-the-art normalised MSE on both the DeepREM and RadioMapSeer benchmarks and, crucially, remain stable when extrapolating to megapixel-scale, irregularly shaped regions. Environment-model-assisted frameworks constitute another sampling-free branch. The spatial loss-field model views shadowing as an integral of incremental loss along weighted propagation ellipses and solves a linear inverse problem to recover the loss field from sparse node-pair observations \cite{li2022radionet,wang2024radiodiff,li2024radiogat,li2024radiogat}. In contrast, the virtual-obstacle model replaces real geometry with electromagnetic “obstacle classes”, each characterised by a penetrability parameter and learnable position; optimisation then jointly refines obstacle placement and path-loss parameters through non-linear least squares or CNN surrogates \cite{172,173,174}. These models exhibit favourable extrapolation to 3-D transmitter–receiver pairs, but their non-convex objectives and reliance on obstacle-class heuristics limit scalability.

A unifying observation across the above literature is that most published work remains confined to a single, fixed-height plane and concentrates exclusively on large-scale path loss. Critical 3-D descriptors such as DoA, ToA, delay spread, and angular spread, indispensable for beamforming, interference coordination, and centimetre-level localisation—are almost entirely absent from current RM predictors \cite{149,152}. Architecturally, prevailing CNN and transformer backbones employ 2-D convolution kernels or 2-D axial attention, which are mathematically incapable of capturing volumetric propagation patterns intrinsic to high-rise and aerial networks. The scarcity of public, high-resolution 3-D datasets with co-registered geometry, material metadata, and rich channel labels further compounds the difficulty of training 3-D-native models. Consequently, the emerging consensus is that deep generative learning offers the most plausible route to low-cost, scalable RM construction, provided two bottlenecks are addressed. First, neural operators must be redesigned to process 3-D spatial relationships natively, e.g., through sparse 3-D convolutions, point-cloud transformers, or volumetric diffusion kernels. Second, the community must curate and release high-precision 3-D channel datasets that span multiple frequency bands, polarisation states, and transmitter heights. The framework proposed in this paper tackles both challenges by introducing 3-D diffusion primitives and demonstrating their efficacy on a newly compiled, millimetre-wave 3-D dataset, thereby establishing a foundation for truly environment-aware, inference-efficient 6G systems.

\subsection{Data Acquisition of RM}
Publicly available radio map and channel datasets serve as critical foundations for a wide array of learning-based research in wireless communications. Early developments in this domain primarily leveraged synthetic ray-tracing techniques to circumvent the prohibitive cost of large-scale data collection \cite{zeng2024tutorial}. A representative example is DeepMIMO, which provides a scalable \cite{alkhateeb2019deepmimo}, parametric dataset generator that exports sub-6 GHz, 28 GHz, and 60 GHz MIMO channel matrices across millions of user positions in diverse urban scenarios. These synthetic corpora deliver unmatched environmental precision and scale, offering perfect ground truth for environmental geometry and material properties. However, they are intrinsically constrained by the assumptions embedded within their underlying ray-tracing engines and inherently lack the hardware-induced imperfections and dynamic effects encountered in real-world deployments.

A widely adopted approach for synthesising RMs relies on analytical channel models, wherein the large-scale path-loss surface is obtained by directly evaluating closed-form propagation expressions. In practice, most studies invoke the log-normal shadow-fading formulae prescribed in standardised guidelines—most notably the 3GPP Urban Micro (UMi) and Urban Macro (UMa) models, whose deterministic component expresses the mean path loss as a function of transmitter–receiver (TX–RX) separation, while a zero-mean log-normal term captures the shadowing variance \cite{docomo20165g}. After the geographical coordinates and configuration parameters of all BSs, including height, radiated power, antenna pattern, and carrier frequency, have been specified, the analytical model can be evaluated on a dense spatial grid to yield a path-loss field that is subsequently stored as an RM. Recent learning-based studies employ these analytically generated datasets to pre-train, fine-tune, or supervise neural RM generators \cite{17,18,19}. Although analytically generated RMs are computationally attractive, their fidelity is limited when transplanted to real deployments. The underlying logarithmic-distance formula inherently neglects location-specific propagation mechanisms such as street-canyon diffraction, irregular rooftop scattering, foliage attenuation, and material-dependent penetration loss. These omissions induce non-negligible discrepancies—often exceeding 10 dB—between the synthetic path loss and field measurements, thereby undermining the spatial consistency required for downstream tasks such as beam management, centimetre-level localisation, and proactive interference coordination. To mitigate this deficiency, recent research has sought to inject environmental knowledge, either via explicit electromagnetic simulations or data-driven corrections, into the RM synthesis pipeline. In particular, approximate physical solvers based on deterministic ray tracing have gained traction \cite{mbugua2020review}; by approximately solving Maxwell’s equations over detailed geometric models, ray tracing yields a richer distribution of channel descriptors that better aligns with empirical observations, albeit at a markedly higher computational cost \cite{deschamps1972ray,liebmann1950field}.

To address this realism gap, recent large-scale measurement-driven datasets have been introduced. RadioGAT introduced a hybrid framework combining a log-distance pathloss model with a graph attention network (GAT), enabling the reconstruction of multi-band RMs from sparse measurements. Its dataset covers ten urban subregions ray-traced over five carrier frequencies, demonstrating strong performance under limited supervision \cite{li2024radiogat}. However, its radio maps are restricted to two-dimensional layouts where both transmitters and receivers share the same height. This constraint neglects critical 3D phenomena, such as height-varying signal propagation, near-field effects in XL-MIMO, and vertical non-stationarity—thereby limiting its relevance to elevation-sensitive applications like UAV-assisted communications \cite{wang2024tutorial}. Efforts to incorporate more geometric realism can be seen in the RMDirectionalBerlin dataset, which leverages LiDAR-derived building height profiles and directional rooftop antennas to better reflect true deployment conditions \cite{jaensch2024radio}. With more than 74,000 maps, it supports geometry-free reasoning and vision-based learning using accompanying aerial imagery. However, it lacks CIR-level granularity that is no DoA, ToA, or multipath richness is preserved, and is confined to a single city and on 3.5GHz, precluding analysis of geographical or spectral generalization. The RadioMapSeer dataset provides broader urban diversity, with 56,000 RM samples simulated across six European cities using both low-fidelity and high-fidelity ray tracing methods \cite{levie2021radiounet}. Its inclusion of paired simulations makes it particularly suitable for transfer learning and model distillation studies. Nonetheless, key limitations persist that all buildings are modeled with uniform 25-meter heights, vegetation and dynamic scatterers are ignored, and the dataset supports only one frequency without CSI-level labels, constraining its use for 3D-aware or delay-sensitive modeling. In terms of scale, CKMImageNet presents a hierarchical storage structure that couples large-scale link measurements with 64×64 pixel heatmap visualizations and corresponding environmental image tiles. It enables fine-grained querying and multi-base station scenarios, such as those simulated in Beijing with 42 transmitters and 500,000 users \cite{wu2024ckmimagenet}. While it is a pioneering effort in linking visual and physical domains, many scenes lack explicit DoA/DoD or ToA annotations, and most tiles are simulated using a single BS layout, limiting the diversity of interference topologies. To enable large-scale multi-domain analysis, SpectrumNet \cite{zhang2024generative} offers one of the most comprehensive image-based datasets to date, spanning over 300,000 radio maps across eleven terrain types, five frequency bands, three receiver altitudes, which are 1.5 m, 30 m, and 200 m, and synthetic climate profiles. It is among the first to support altitude-aware CKM and non-terrestrial link modeling. However, its spatial granularity is coarse, restricted to a few discrete heights, and it only provides scalar pathloss values, omitting directional and temporal channel features that are vital for beamforming, localization, and channel reconstruction in high-mobility 6G settings.

\subsection{Generative Diffusion Model}
Recent advances in generative modeling have seen diffusion models emerge as a promising alternative to traditional frameworks, such as GAN \cite{creswell2018generative} and variational autoencoder (VAE) \cite{kingma2013auto}, offering improved training stability and high-quality sample generation \cite{kang2025confidence,11004012,10990238,liu2025optimizing,zhao2024signal,zhao2024signal}. Different from GANs, which rely on adversarial objectives and often suffer from training instability, diffusion models leverage a likelihood-based formulation that progressively refines samples from noise to data through a denoising process \cite{croitoru2023diffusion}. Among these, the denoising diffusion probabilistic model (DDPM) has demonstrated remarkable success in diverse domains, including computer vision, natural language processing, and reinforcement learning \cite{ho2020denoising}. DDPM models the data generation process as a two-stage Markov chain comprising a forward diffusion process and a reverse denoising process. In the forward process, Gaussian noise is incrementally added to clean data \(\bm{x}_0\), resulting in a sequence \(\bm{x}_1, \bm{x}_2, \ldots, \bm{x}_T\), defined as follows.
\begin{equation}
q(\bm{x}_1,\ldots,\bm{x}_T\mid\bm{x}_0)=\prod_{t=1}^Tq(\bm{x}_t\mid\bm{x}_{t-1}),
\end{equation}
\begin{equation}
q(\bm{x}_t\mid \bm{x}_{t-1}) = \mathcal{N}(\sqrt{1-\beta_t} \bm{x}_{t-1}, \beta_t \bm{I}),
\end{equation}
where \(\beta_t\) is a small noise variance scheduled over time. Using the cumulative product \(\bar{\alpha}_t = \prod_{s=1}^{t} (1 - \beta_s)\), the distribution of \(\bm{x}_t\) conditioned on \(\bm{x}_0\) is given in closed form as follows.
\begin{equation}
q(\bm{x}_t\mid\bm{x}_0)=\mathcal{N}(\sqrt{\bar{\alpha}_t} \bm{x}_0, (1 - \bar{\alpha}_t) \bm{I}),
\end{equation}
\begin{equation}
\bm{x}_t = \sqrt{\bar{\alpha}_t} \bm{x}_0 + \sqrt{1 - \bar{\alpha}_t} \bm{\epsilon},\quad \bm{\epsilon} \sim \mathcal{N}(0, \bm{I}).
\end{equation}

The reverse process attempts to reconstruct the original data by learning the reverse transition distribution as follows.
\begin{equation}
p_\theta(\bm{x}_{t-1} \mid \bm{x}_t) = \mathcal{N}(\bm{\mu}_\theta(\bm{x}_t, t), \beta_t \bm{I}),
\end{equation}
\begin{equation}
\bm{x}_{t-1} = \frac{1}{\sqrt{\alpha_t}} \left( \bm{x}_t - \frac{1 - \alpha_t}{\sqrt{1 - \bar{\alpha}_t}} \bm{\mu}_\theta(\bm{x}_t, t) \right) + \beta_t \bm{I}.
\end{equation}
While the theoretically correct variance scaling factor is \(\frac{1-\bar{\alpha}_{t-1}}{1-\bar{\alpha}_t} \beta_t\), it has been shown in \cite{ho2020denoising} that using \(\beta_t\) simplifies implementation without degrading performance.

Despite the fidelity achieved by DDPM, its reliance on a large number of denoising steps significantly impacts inference latency. To alleviate this, the denoising diffusion implicit model (DDIM) has been introduced as a deterministic and non-Markovian alternative \cite{song2020denoising}. DDIM maintains compatibility with the DDPM training objective but redefines the reverse process to allow for faster, deterministic sampling as follows.
\begin{equation}
\bm{x}_t = \sqrt{\bar{\alpha}_t} \bm{x}_0 + \sqrt{1 - \bar{\alpha}_t} \bm{\epsilon}, \quad \bm{\epsilon} \sim \mathcal{N}(0, \bm{I}),
\end{equation}
\begin{equation}
\bm{\epsilon}_t = \frac{\bm{x}_t - \sqrt{\bar{\alpha}_t} \bm{x}_0}{\sqrt{1 - \bar{\alpha}_t}}, 
\end{equation}
\begin{equation}
\bm{x}_{t-1} = \sqrt{\bar{\alpha}_{t-1}} \bm{x}_0 + \sqrt{1 - \bar{\alpha}_{t-1}} \bm{\epsilon}_t.
\end{equation}
Furthermore, DDIM introduces a hyperparameter \(\eta \in [0, 1]\) to control the degree of stochasticity in the sampling path. When \(\eta = 0\), the trajectory becomes fully deterministic, leading to consistent sample generation with far fewer steps. This property is particularly beneficial for time-sensitive applications, such as real-time 3D radio map construction in highly dynamic 6G environments, where low latency and reliability are essential. By leveraging the strengths of both DDPM and DDIM, the proposed generative framework in this work inherits high-quality generation capabilities while enabling efficient inference suitable for practical deployment in wireless systems.

\section{Properties of UrbanRadio3D}

\subsection{3D RadioMap Dataset Construction}
In this section, the construction of the UrbanRadio3D dataset is detailed. The dataset has been developed using the WinProp module in the Altair software suite to simulate three-dimensional radio frequency propagation in realistic urban environments. It comprises 701 distinct urban regions, each modeled as a $256 \times 256\,\text{m}^2$ area at a spatial resolution of 1 meter per pixel. These regions are derived from representative global cities, including Kara, Berlin, Glasgow, Ljubljana, London, and Tel Aviv, capturing diverse architectural and geographical characteristics. To enhance realism, the simulation accounts for varying building heights across cities, ranging from 6.6 m to 19.8 m. For each urban map, 200 transmitter locations were randomly selected, and simulations were conducted at receiver heights from 1 m to 20 m in 1-meter increments. This setup yields a total of $701 \times 200 \times 20 = 2.84$ million simulation instances, resulting in a large-scale dataset that accurately reflects volumetric wireless propagation in urban scenarios. Each simulation result is stored as a high-resolution PNG image, maintaining 1 m spatial precision across three dimensions. The dataset includes multiple signal modalities, such as pathloss, DoA, and ToA, along with supplementary representations to support advanced modeling. These include polygonal masks of building structures, grayscale elevation maps encoding building heights, and transmitter location maps. Notably, both DPM-based simulations \cite{wahl2005dominant} and auxiliary channel metrics are incorporated to ensure physical interpretability and modeling consistency. Key simulation parameters are summarized in Table~\ref {tab:additional_parameters}. Each transmitter emits at 23 dBm/Hz using an isotropic antenna operating at a carrier frequency of 5.9 GHz, representative of mmWave scenarios in future 6G systems. The receiver height is nominally set at 1.5 m, with measurements collected over the vertical range of 1–20 m to facilitate full 3D modeling. Thresholding is applied to the pathloss, DoA, and ToA components to suppress physically implausible outliers and ensure model training focuses on meaningful signal interactions. All data have been generated using the Dominant Path Model (DPM) \cite{wahl2005dominant}, providing a balance between computational efficiency and physical accuracy. \added{To construct a volumetric 3D radio map, we utilized the electromagnetic simulator FEKO, which performs electromagnetic field rendering on 2D planes at fixed receiver heights. Since each simulation run returns results only for a specific height, we conducted separate simulations at 20 different height levels, ranging from ground to rooftop level. This procedure effectively yields 20 horizontal slices per urban scene, which are then stacked to form a complete 3D spatial distribution of wireless channel characteristics. The spatial resolution within each 2D slice was set to 1 meter, consistent with the settings used in RadioMapSeer \cite{levie2021radiounet}, to ensure reproducibility and fair comparisons with existing baselines.}

\added{For the simulation of material properties, the default material parameters provided by the WinProp software were used. To simplify the modeling process, all building facades in the simulation were assigned the same default material properties, as summarized in Table \ref{tab:material_defaults}. This approach assumes uniform electromagnetic behavior across all building surfaces, which facilitates large-scale simulation while maintaining reasonable physical accuracy.}


\begin{figure}[h]
\centering
\setlength\fboxsep{0pt} 
\subfloat[]{\fbox{\includegraphics[width=0.13\textwidth]{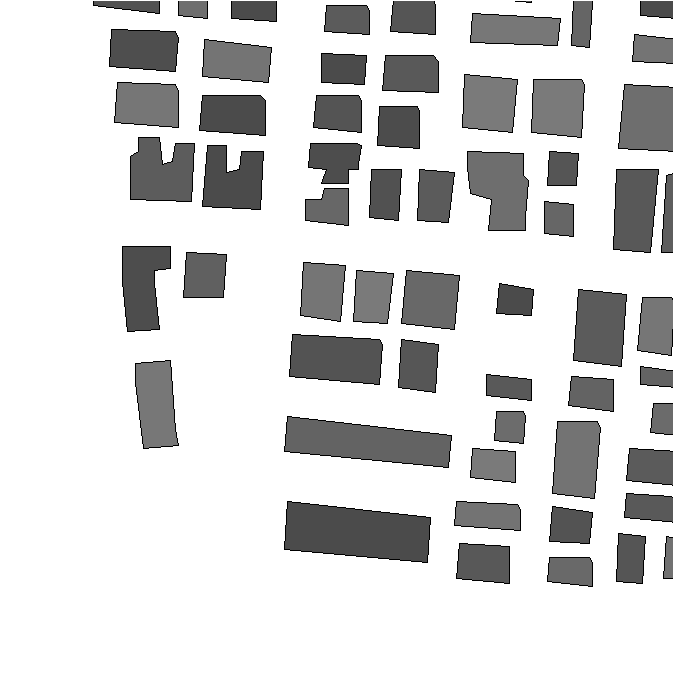}}}\hspace{0.5em}
\subfloat[]{\fbox{\includegraphics[width=0.13\textwidth]{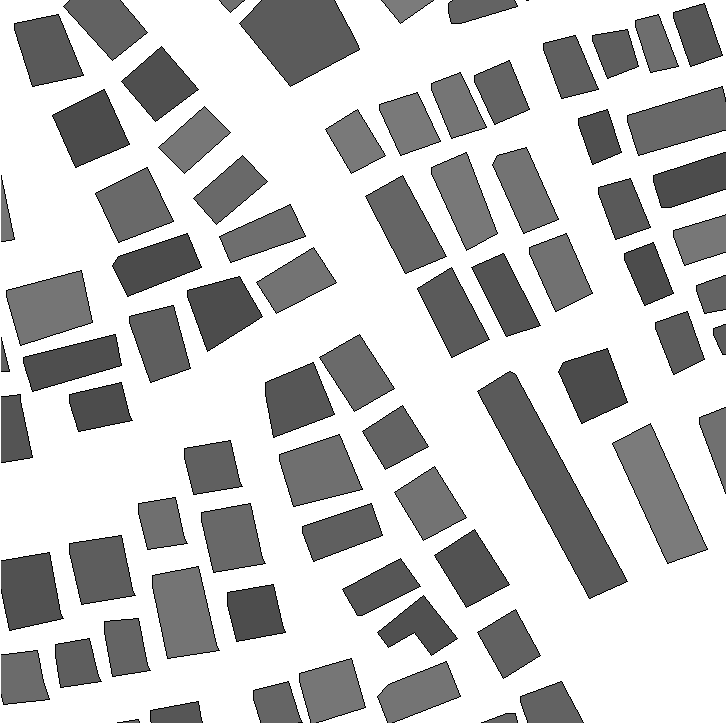}}}\hspace{0.5em}
\subfloat[]{\fbox{\includegraphics[width=0.13\textwidth]{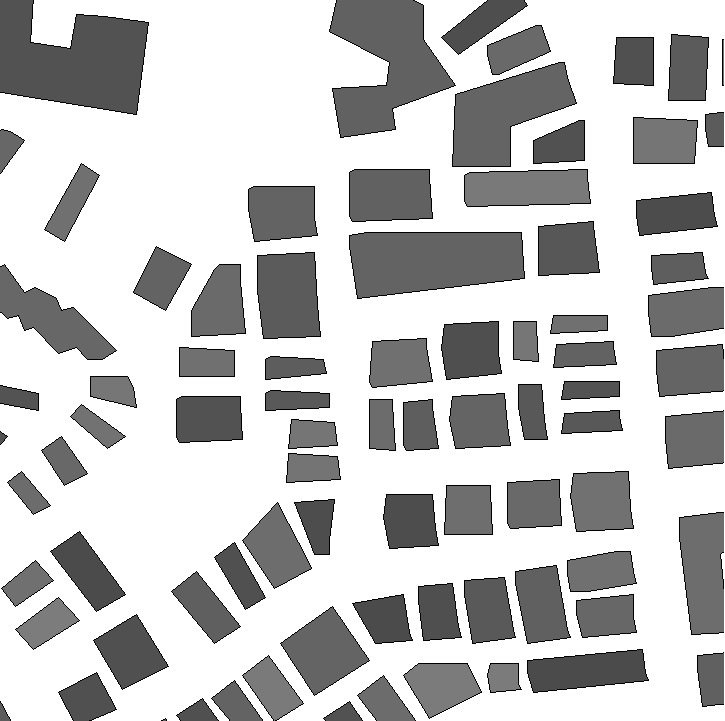}}}\\[0.1em]
\subfloat[]{\fbox{\includegraphics[width=0.13\textwidth]{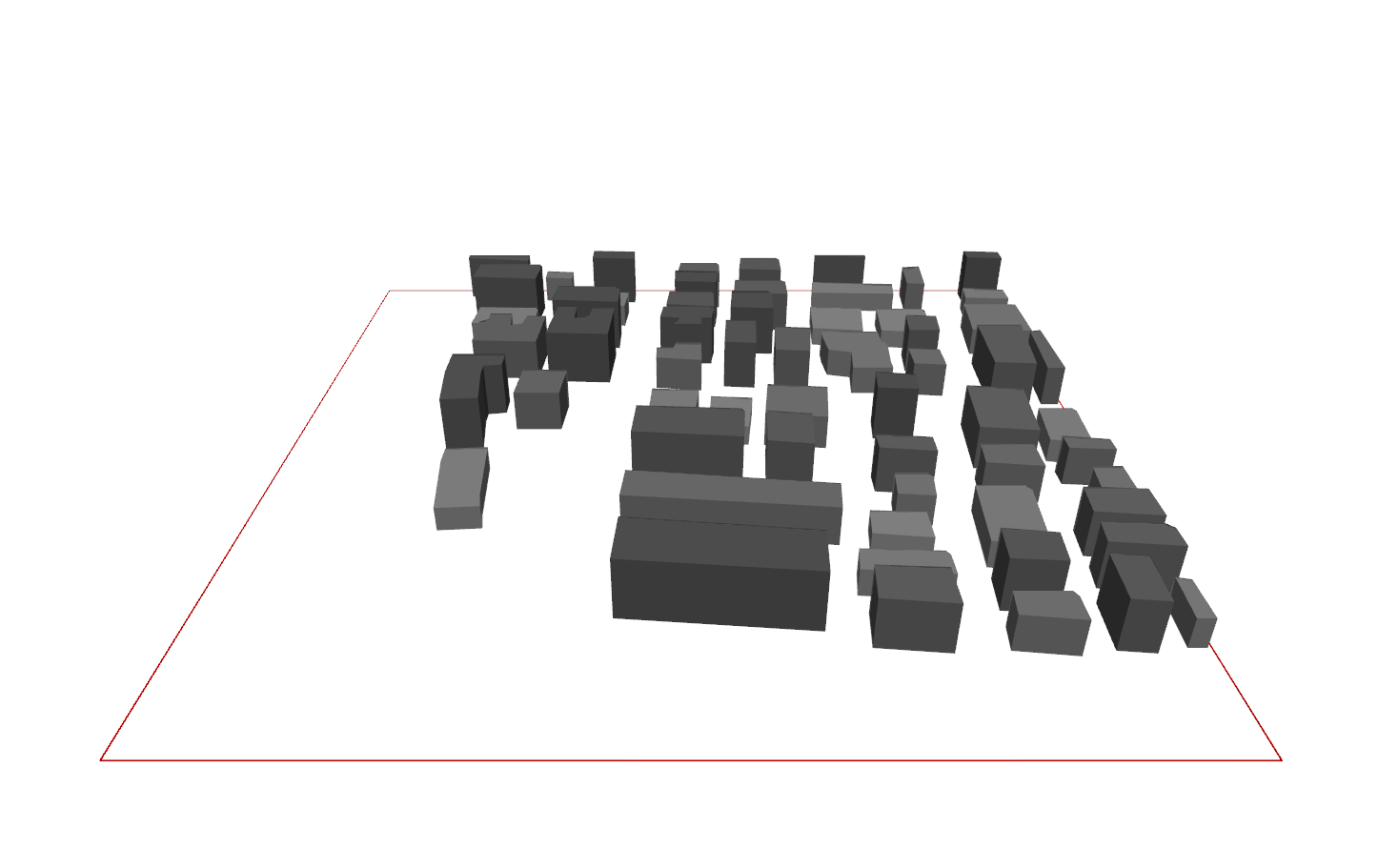}}}\hspace{0.5em}
\subfloat[]{\fbox{\includegraphics[width=0.13\textwidth]{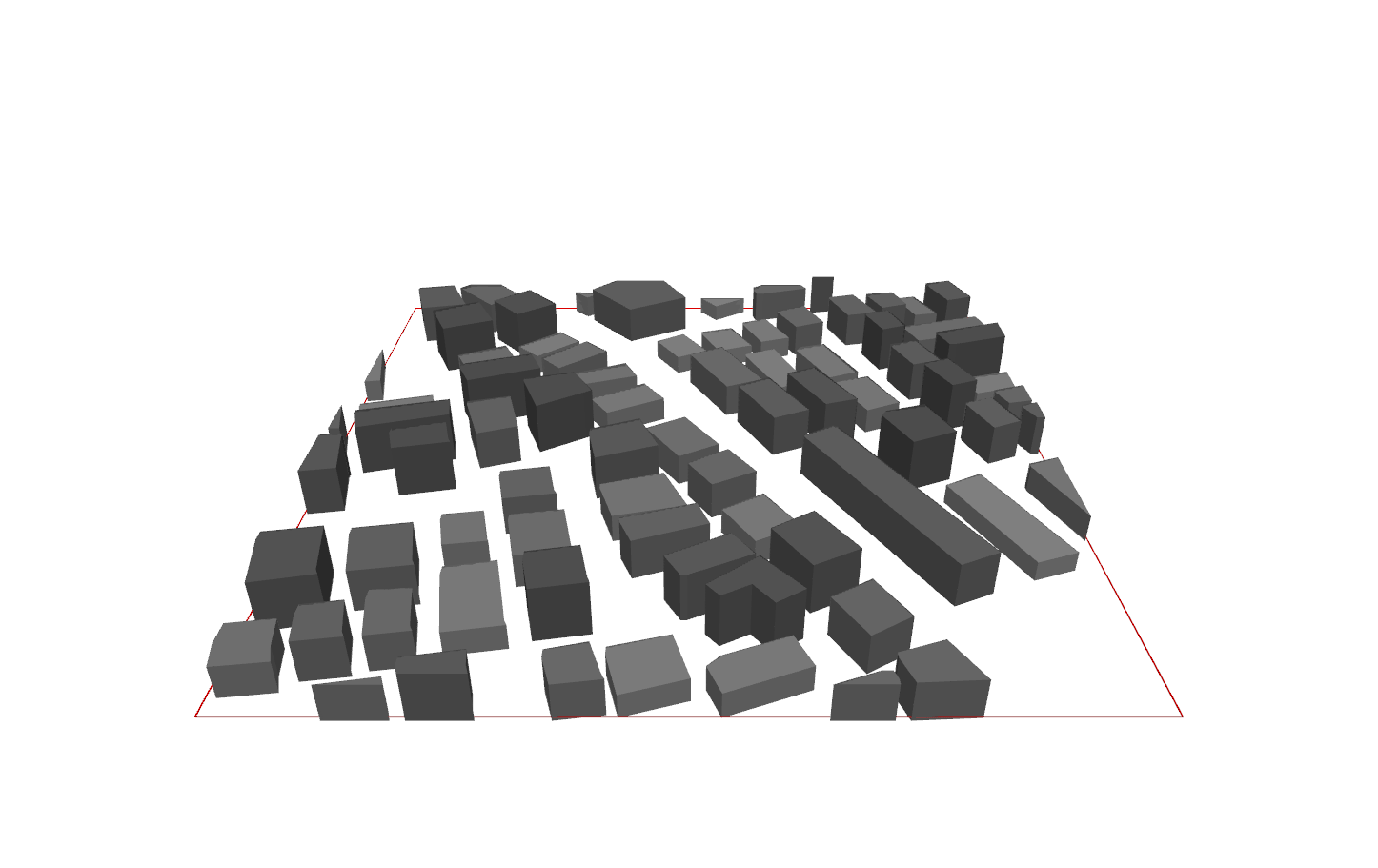}}}\hspace{0.5em}
\subfloat[]{\fbox{\includegraphics[width=0.13\textwidth]{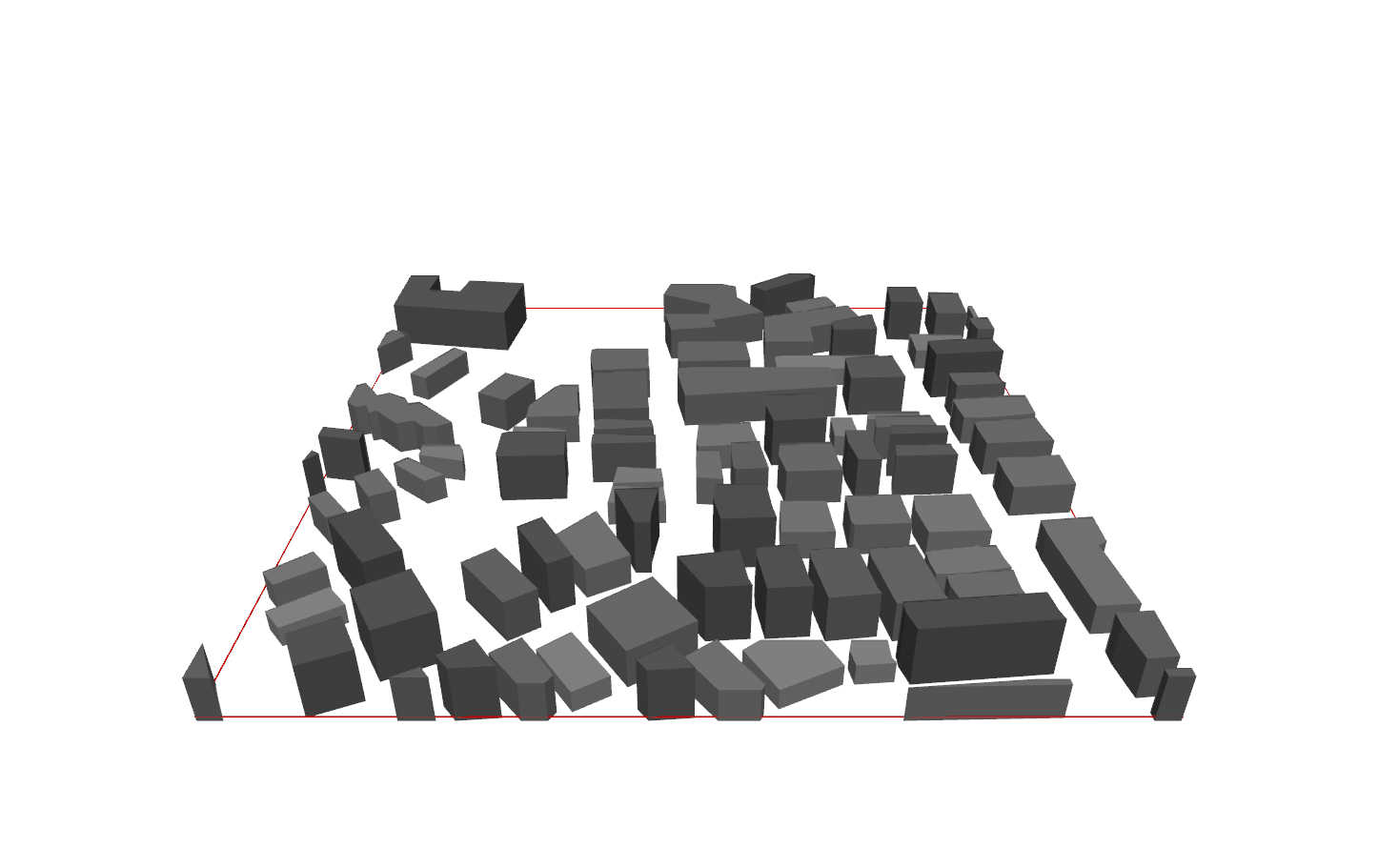}}}
\caption{2D views (top) and 3D views (bottom) of three scenarios.}
\end{figure}

\begin{figure}[h]
\centering
\setlength\fboxsep{0pt} 
\subfloat[]{\fbox{\includegraphics[width=0.15\textwidth]{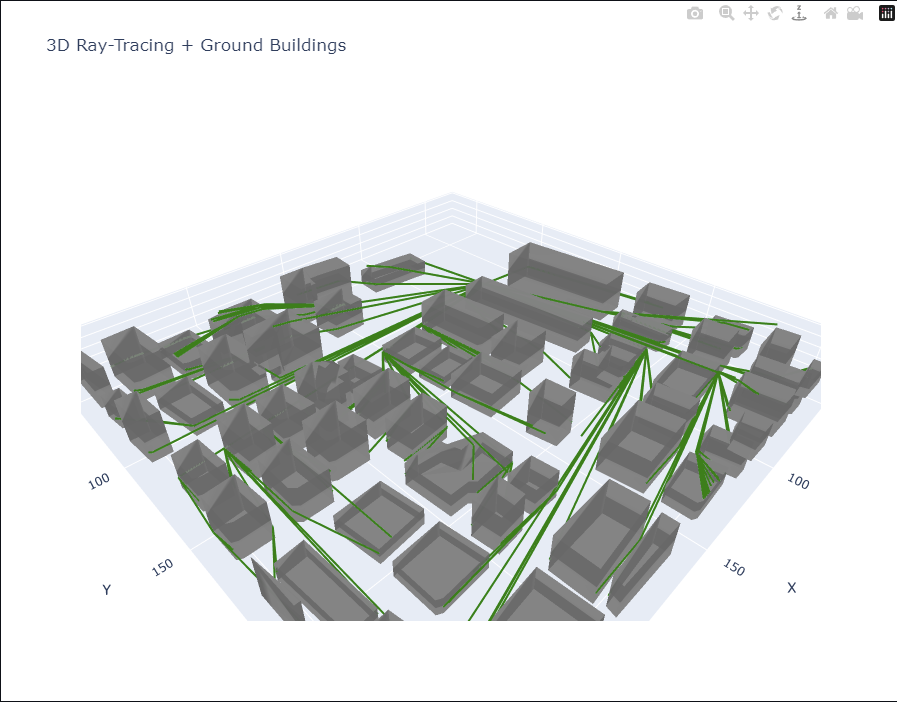}}}\hspace{0.1em}
\subfloat[]{\fbox{\includegraphics[width=0.15\textwidth]{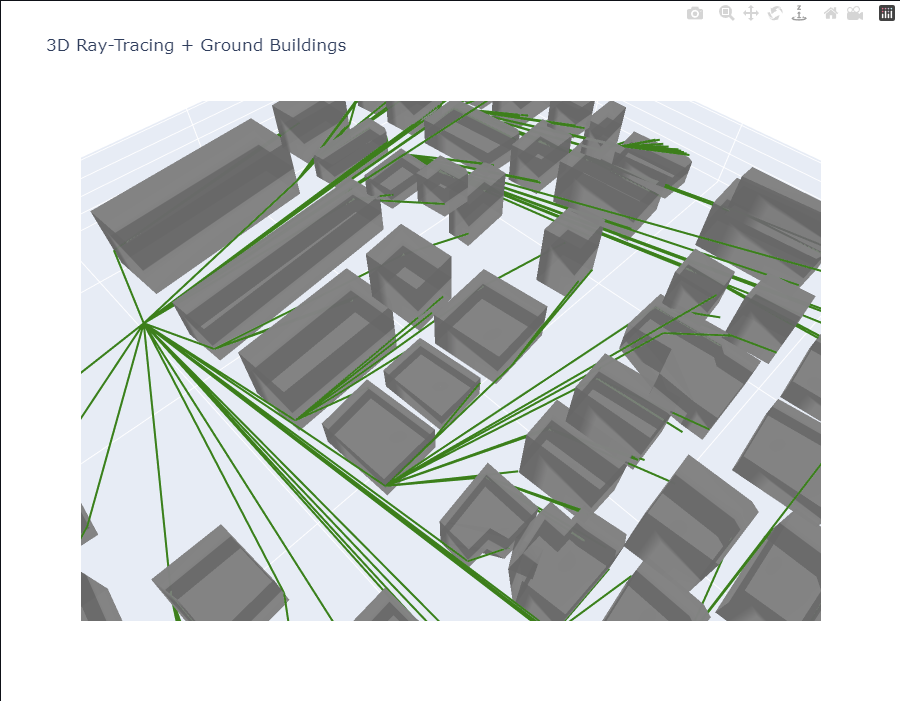}}}\hspace{0.1em}
\subfloat[]{\fbox{\includegraphics[width=0.15\textwidth]{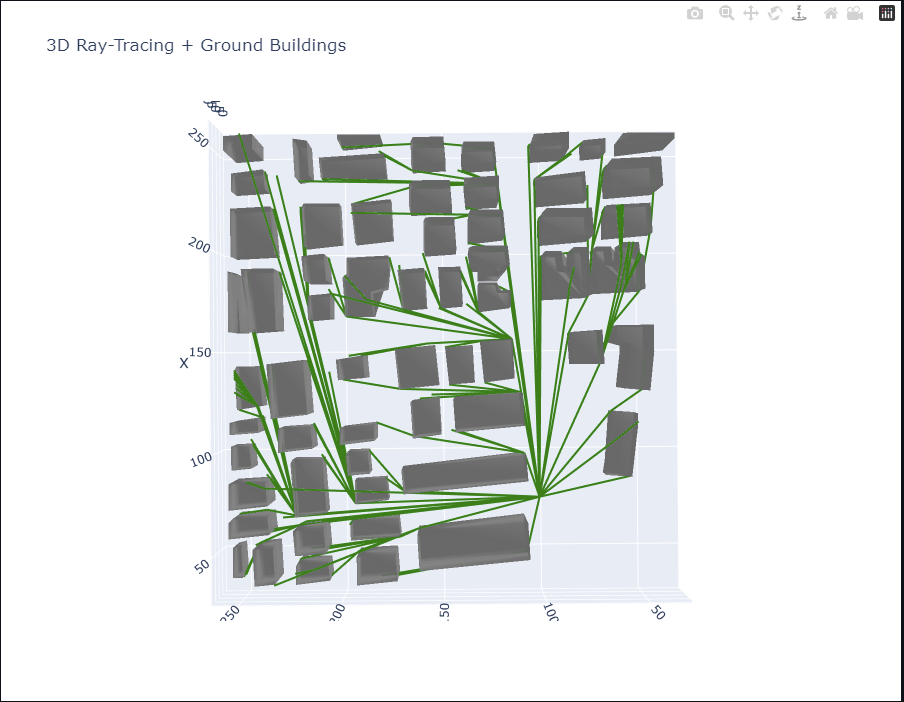}}}\\
\caption{3D ray-tracing views}
\label{fig:ray}
\end{figure}

\FloatBarrier

\begin{figure}[h]
\centering
\subfloat[Building Segmentation Map]{\includegraphics[width=0.14\textwidth]{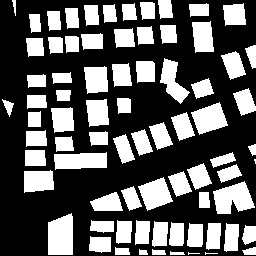}}\hspace{0.5em}
\subfloat[Building Height Map]{\includegraphics[width=0.14\textwidth]{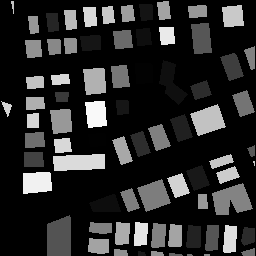}}\hspace{0.5em}
\subfloat[Transmitter Location Map]{\includegraphics[width=0.14\textwidth]{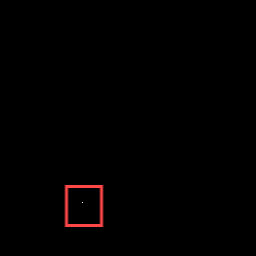}}\hspace{0.5em}
\caption{Examples of structural and spatial information maps included in the dataset.}
\label{fig:urban}
\end{figure}

\begin{table}[ht]
\caption{Additional Parameters of the 3D RadioMap Dataset}
\centering
\renewcommand{\arraystretch}{1.4}
\begin{tabular}{@{}cc|c@{}}
\toprule
\textbf{Parameter} & & \textbf{Value} \\
\midrule
Map size & & $256 \times 256\,\text{m}^2$ \\
Pixel length & & $1\,\text{m}$ \\
Rx height & & $1.5\,\text{m}$ \\
Observation height & & $1.0$--$20.0\,\text{m}$ \\
Height range of the buildings & & $6.6$--$19.8\,\text{m}$ \\
Transmit power & & $23\,\text{dBm/Hz}$ \\
Antenna type & & Isotropic \\
Center carrier frequency & & $5.9\,\text{GHz}$ \\
PL Min threshold & & $-169.0\,\text{dB}$ \\
PL Max threshold & & $-92.0\,\text{dB}$ \\
ToA Min threshold & & $0.0\,\text{ns}$ \\
ToA Max threshold & & $1180.0\,\text{ns}$ \\
DoA Azi Min threshold & & $0.0\,\text{Rad}$ \\
DoA Azi Max threshold & & $6.3\,\text{Rad}$ \\
DoA Ele Min threshold & & $0.5\,\text{Rad}$ \\
DoA Ele Max threshold & & $2.25\,\text{Rad}$ \\
Simulation type & & DPM \\
\bottomrule
\end{tabular}
\label{tab:additional_parameters}
\end{table}

\begin{table}[ht]
\caption{\added{Default Material Parameters Used in Simulation}}
\centering
\renewcommand{\arraystretch}{1.4}
\begin{tabular}{@{}cc|c@{}}
\toprule
\added{\textbf{Parameter}} & & \added{\textbf{Value}} \\
\midrule
\added{Surface thickness} & & \added{$10.0\,\text{cm}$} \\
\added{Surface roughness} & & \added{$0.0000\,\mu\text{m}$} \\
\added{Operating frequency} & & \added{$2000\,\text{MHz}$} \\
\added{Material selection rule} & & \added{Nearest frequency used} \\
\added{Transmission loss} & & \added{$20.0\,\text{dB}$} \\
\added{Reflection loss} & & \added{$9.0\,\text{dB}$} \\
\added{Diffraction incident (min)} & & \added{$8.0\,\text{dB}$} \\
\added{Diffraction incident (max)} & & \added{$15.0\,\text{dB}$} \\
\added{Diffracted loss} & & \added{$5.0\,\text{dB}$} \\
\added{Relative permittivity $\varepsilon_r$} & & \added{$4.0$} \\
\added{Relative permeability $\mu_r$} & & \added{$1.0$} \\
\added{Electrical conductivity $\sigma$} & & \added{$0.01\,\text{S/m}$} \\
\added{Fresnel coefficients} & & \added{Enabled} \\
\added{GTD/UTD diffraction} & & \added{Enabled} \\
\bottomrule
\end{tabular}
\label{tab:material_defaults}
\end{table}



\subsection{Visualization of the Dataset}

\begin{figure*}[t]
\centering
\setlength\fboxsep{0pt}        
\setlength\fboxrule{0.5pt}     

\begin{minipage}{0.18\textwidth}
  \fbox{\includegraphics[width=\linewidth]{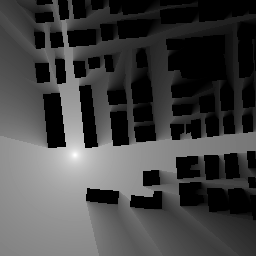}}
  \caption*{\footnotesize \text{0\_X101\_Y3 @ 3\,m}}
\end{minipage}\hspace{0.5em}
\begin{minipage}{0.18\textwidth}
  \fbox{\includegraphics[width=\linewidth]{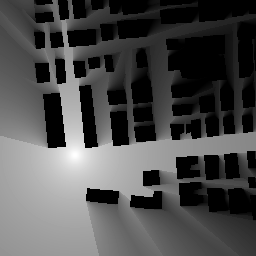}}
  \caption*{\footnotesize \text{0\_X101\_Y3 @ 6\,m}}
\end{minipage}\hspace{0.5em}
\begin{minipage}{0.18\textwidth}
  \fbox{\includegraphics[width=\linewidth]{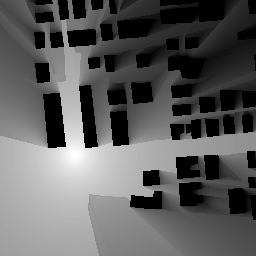}}
  \caption*{\footnotesize \text{0\_X101\_Y3 @ 9\,m}}
\end{minipage}\hspace{0.5em}
\begin{minipage}{0.18\textwidth}
  \fbox{\includegraphics[width=\linewidth]{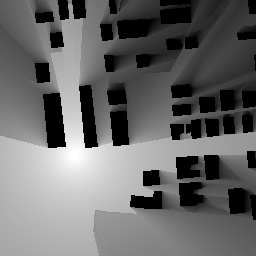}}
  \caption*{\footnotesize \text{0\_X101\_Y3 @ 12\,m}}
\end{minipage}\hspace{0.5em}
\begin{minipage}{0.18\textwidth}
  \fbox{\includegraphics[width=\linewidth]{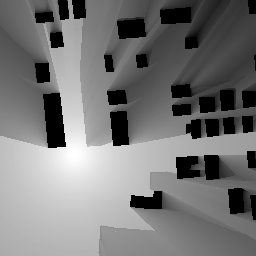}}
  \caption*{\footnotesize \text{0\_X101\_Y3 @ 15\,m}}
\end{minipage}\\[0.8em]

\begin{minipage}{0.18\textwidth}
  \fbox{\includegraphics[width=\linewidth]{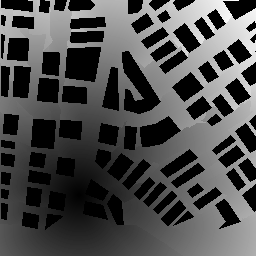}}
  \caption*{\footnotesize \text{5\_X79\_Y61 @ 3\,m}}
\end{minipage}\hspace{0.5em}
\begin{minipage}{0.18\textwidth}
  \fbox{\includegraphics[width=\linewidth]{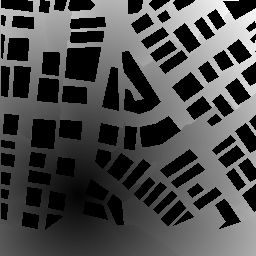}}
  \caption*{\footnotesize \text{5\_X79\_Y61 @ 6\,m}}
\end{minipage}\hspace{0.5em}
\begin{minipage}{0.18\textwidth}
  \fbox{\includegraphics[width=\linewidth]{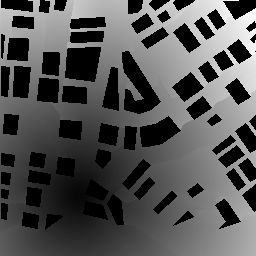}}
  \caption*{\footnotesize \text{5\_X79\_Y61 @ 9\,m}}
\end{minipage}\hspace{0.5em}
\begin{minipage}{0.18\textwidth}
  \fbox{\includegraphics[width=\linewidth]{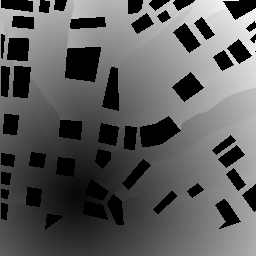}}
  \caption*{\footnotesize \text{5\_X79\_Y61 @ 12\,m}}
\end{minipage}\hspace{0.5em}
\begin{minipage}{0.18\textwidth}
  \fbox{\includegraphics[width=\linewidth]{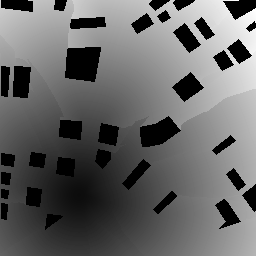}}
  \caption*{\footnotesize \text{5\_X79\_Y61 @ 15\,m}}
\end{minipage}\\[0.8em]

\begin{minipage}{0.18\textwidth}
  \fbox{\includegraphics[width=\linewidth]{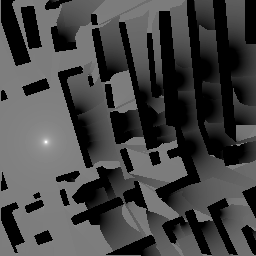}}
  \caption*{\footnotesize \text{296\_X46\_Y114 @ 3\,m}}
\end{minipage}\hspace{0.5em}
\begin{minipage}{0.18\textwidth}
  \fbox{\includegraphics[width=\linewidth]{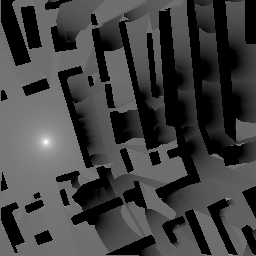}}
  \caption*{\footnotesize \text{296\_X46\_Y114 @ 6\,m}}
\end{minipage}\hspace{0.5em}
\begin{minipage}{0.18\textwidth}
  \fbox{\includegraphics[width=\linewidth]{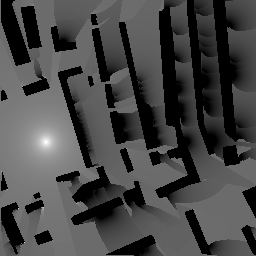}}
  \caption*{\footnotesize \text{296\_X46\_Y114 @ 9\,m}}
\end{minipage}\hspace{0.5em}
\begin{minipage}{0.18\textwidth}
  \fbox{\includegraphics[width=\linewidth]{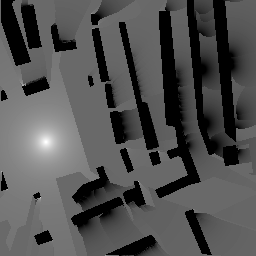}}
  \caption*{\footnotesize \text{296\_X46\_Y114 @ 12\,m}}
\end{minipage}\hspace{0.5em}
\begin{minipage}{0.18\textwidth}
  \fbox{\includegraphics[width=\linewidth]{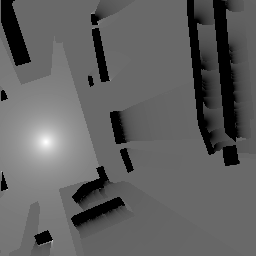}}
  \caption*{\footnotesize \text{296\_X46\_Y114 @ 15\,m}}
\end{minipage}\\[0.8em]

\begin{minipage}{0.18\textwidth}
  \fbox{\includegraphics[width=\linewidth]{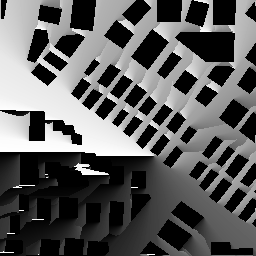}}
  \caption*{\footnotesize \text{37\_X154\_Y100 @ 3\,m}}
\end{minipage}\hspace{0.5em}
\begin{minipage}{0.18\textwidth}
  \fbox{\includegraphics[width=\linewidth]{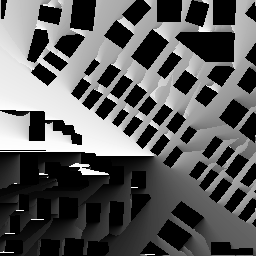}}
  \caption*{\footnotesize \text{37\_X154\_Y100 @ 6\,m}}
\end{minipage}\hspace{0.5em}
\begin{minipage}{0.18\textwidth}
  \fbox{\includegraphics[width=\linewidth]{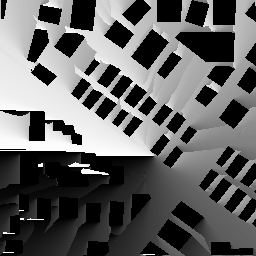}}
  \caption*{\footnotesize \text{37\_X154\_Y100 @ 9\,m}}
\end{minipage}\hspace{0.5em}
\begin{minipage}{0.18\textwidth}
  \fbox{\includegraphics[width=\linewidth]{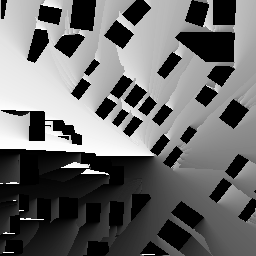}}
  \caption*{\footnotesize \text{37\_X154\_Y100 @ 12\,m}}
\end{minipage}\hspace{0.5em}
\begin{minipage}{0.18\textwidth}
  \fbox{\includegraphics[width=\linewidth]{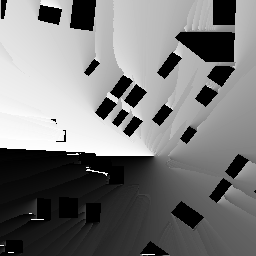}}
  \caption*{\footnotesize \text{37\_X154\_Y100 @ 15\,m}}
\end{minipage}

\vspace{1em}
\caption{%
Visualization of channel characteristics at different observation heights (in meters).
\textbf{Top row:} Pathloss maps at heights 3\,m, 6\,m, 9\,m, 12\,m, and 15\,m, showing the signal attenuation across the environment.
\textbf{Second row:} ToA spread maps at the same heights, capturing the multipath propagation ToAs.
\textbf{Third row:} DoA Elevation angle distributions, indicating the vertical arrival direction of signals.
\textbf{Fourth row:}  DoA Azimuth angle distributions at these heights,
representing the horizontal arrival direction of signals. 
}
\label{fig:characteristics}
\end{figure*}

To provide an intuitive understanding of the dataset, we present several groups of visualizations that demonstrate the diversity and richness of the data. These visualizations include pathloss maps captured at different receiver heights (ranging from 1 meter to 20 meters), which highlight how signal propagation is influenced by changes in elevation and building occlusions. As shown in Fig.~\ref{fig:characteristics}. in addition to pathloss, we also visualize auxiliary channel properties such as ToA, DoA azimuth, DoA elevation, and ray-traced propagation paths(Fig.~\ref{fig:ray}) offering a more comprehensive view of the radio propagation environment. Furthermore, as is shown in Fig.~\ref{fig:urban}, structural and spatial information maps are provided, including building segmentation maps, building height maps, and transmitter location maps. These maps enable a better understanding of the urban spatial context, which is crucial for tasks such as geometric learning, beamforming, and positioning.


Notably, when the observation height exceeds the height of surrounding buildings, these structures may visually “disappear” from the radio map. This occurs because the receiver gains a clear line-of-sight to the transmitter, eliminating obstructions that would otherwise cause signal attenuation or reflection, thus significantly altering the spatial signal patterns.

\subsection{Dataset Naming Conventions}
\begin{figure}[ht]
  \centering
  \subfloat[\SI{1}{m} height slice]{%
    \fbox{\includegraphics[width=0.40\textwidth]{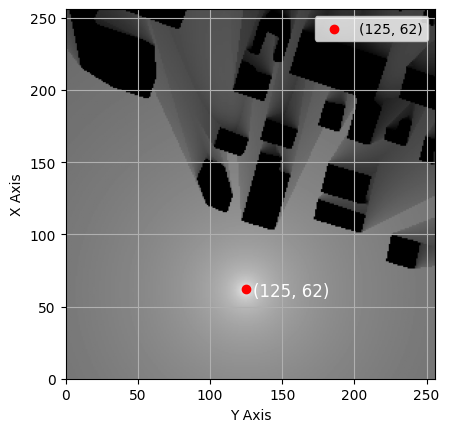}}%
  }
  \caption{Visualisation of the file \texttt{103\_62X\_125Y.png}.
  According to the naming convention, the first field
  (\texttt{103}) denotes the building identifier~(\emph{BID});
  the second field (\texttt{62X}) is the transmitter’s X‑coordinate; and
  the third field (\texttt{125Y}) is the transmitter’s Y‑coordinate
  in the global map.  The image corresponds to the \SI{1}{m} height layer
  and clearly reveals the path‑loss distribution around the transmitter
  and surrounding urban structures.}
  \label{fig:example_slice}
\end{figure}

The dataset used in this study adheres to a standardized folder structure and fil   e-naming convention to ensure consistency and ease of access. At the root level, the dataset is organized into five modality-specific directories: \texttt{pathLoss}, \texttt{Doa\_Azi}, \texttt{Doa\_Ele}, \texttt{ToA}, and \texttt{propagation\_ray}. Each modality folder contains subdirectories labeled as “h1” through “h20”, corresponding to receiver height levels ranging from 1~m to 20~m. Within each height-specific subfolder, data files are named as follows.
\begin{align}
    \underbrace{\textit{BID}}_{\text{Building ID}}\,\_\,\underbrace{\textit{X}}_{\text{X-coordinate}}\,\_X\,\underbrace{\textit{Y}}_{\text{Y-coordinate}}\,\_Y.\{\texttt{png},\,\texttt{npy}\},
\end{align}
where \textit{BID} denotes the building identifier, and \textit{X} and \textit{Y} represent the horizontal coordinates of the transmitter location.

\subsection{Data Normalization and Heatmap Generation}

\added{To ensure consistency across different spatial locations and enable effective visualization and model training, we adopt a \textbf{global normalization} strategy for all channel parameters, including pathloss (PL), time of arrival (ToA or Delay), direction of arrival azimuth (DoA\_Azi), and direction of arrival elevation (DoA\_Ele).}

\added{Each parameter is normalized individually using fixed global minimum and maximum thresholds as specified in Table~\ref{tab:additional_parameters}. The normalization formula is defined as:}

\begin{equation}
\added{
x_{\text{norm}} = \max\left(0, \frac{x - x_{\min}}{x_{\max} - x_{\min}}\right)
}
\label{eq:normalization}
\end{equation}

\added{Here, $x$ denotes the original parameter value, $x_{\min}$ and $x_{\max}$ are the predefined global minimum and maximum thresholds, and $x_{\text{norm}} \in [0, 1]$ is the normalized result. For visualization or model input purposes, the normalized values can be further mapped to the range $[0, 255]$  as needed.}

\begin{figure*}[t]
\centering
\setlength\fboxsep{0pt}
\setlength\fboxrule{0.5pt}

\begin{minipage}{0.22\textwidth}
  \fbox{\includegraphics[width=\linewidth]{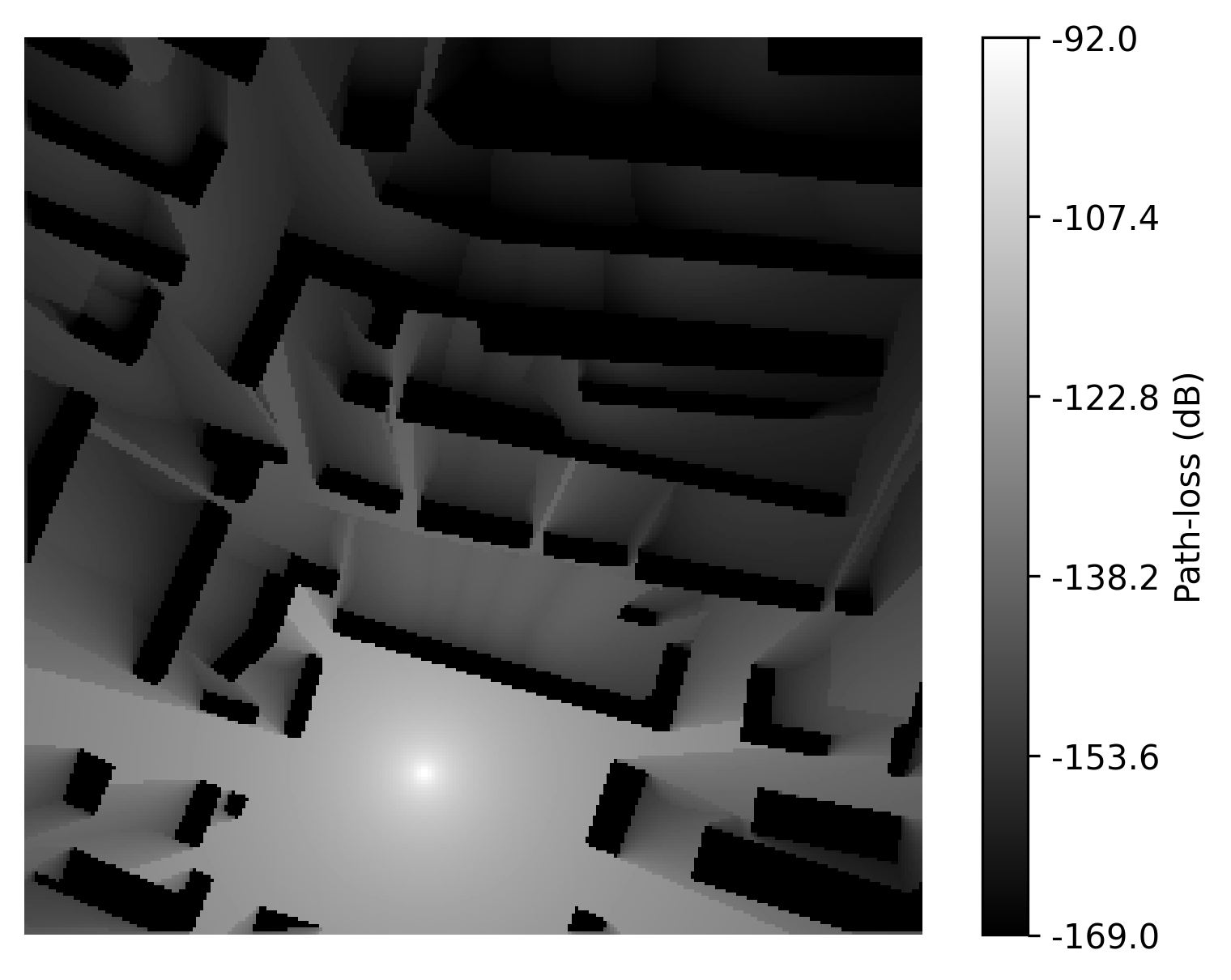}}
  \par\centering
  {\footnotesize \added{Pathloss maps}}
\end{minipage}\hspace{0.5em}
\begin{minipage}{0.22\textwidth}
  \fbox{\includegraphics[width=\linewidth]{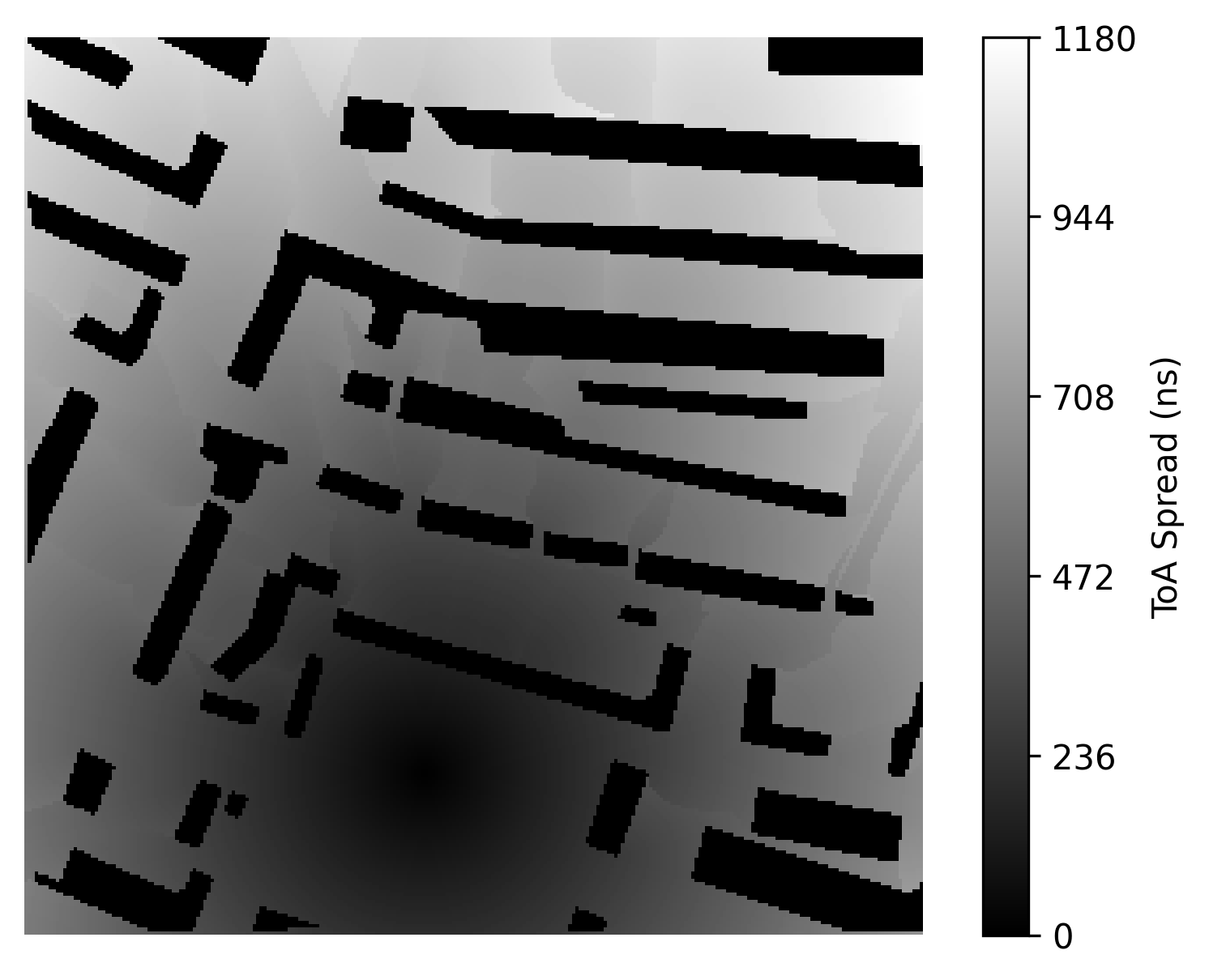}}
  \par\centering
  {\footnotesize \added{ToA spread}}
\end{minipage}\hspace{0.5em}
\begin{minipage}{0.22\textwidth}
  \fbox{\includegraphics[width=\linewidth]{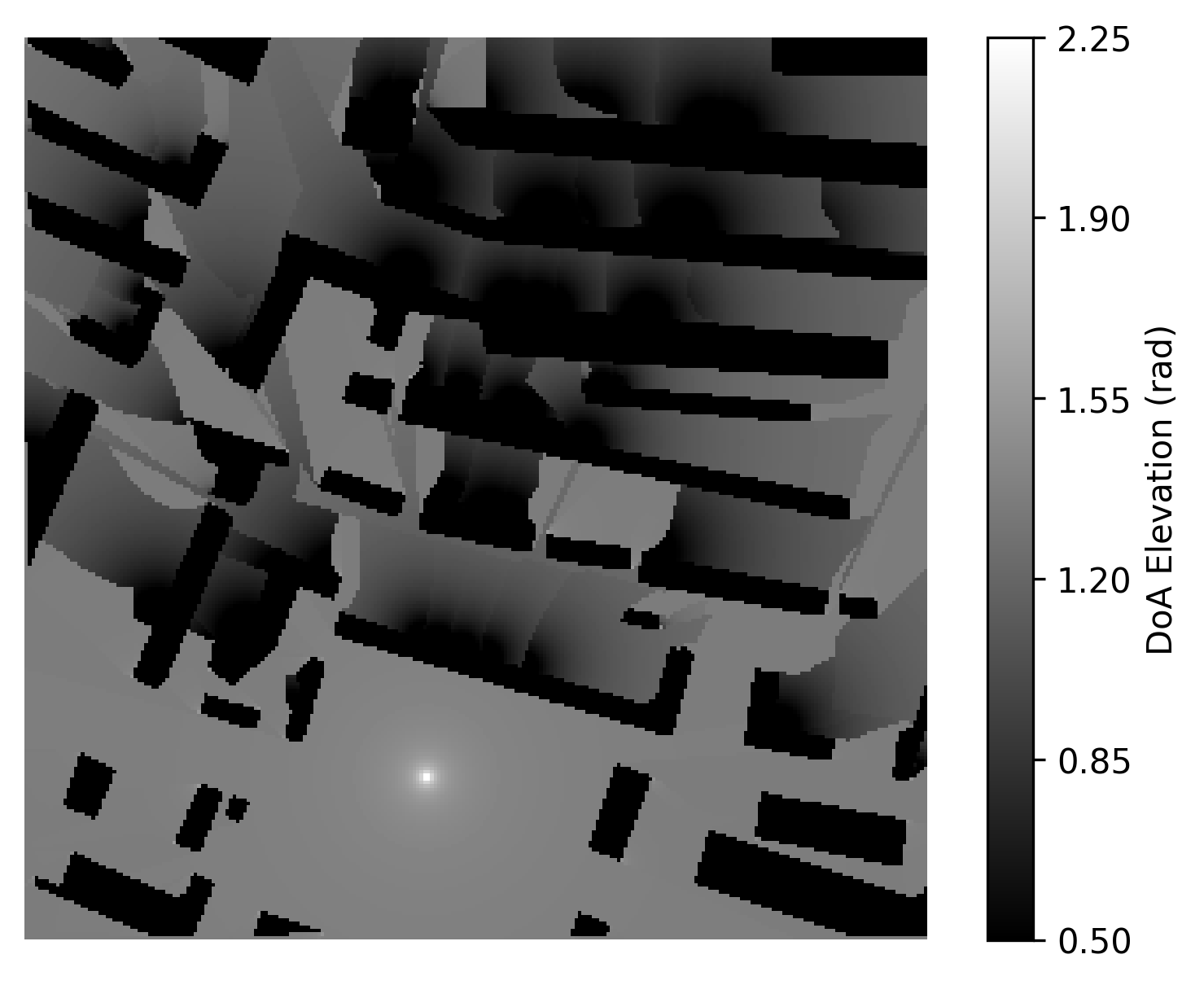}}
  \par\centering
  {\footnotesize \added{DoA Elevation}}
\end{minipage}\hspace{0.5em}
\begin{minipage}{0.22\textwidth}
  \fbox{\includegraphics[width=\linewidth]{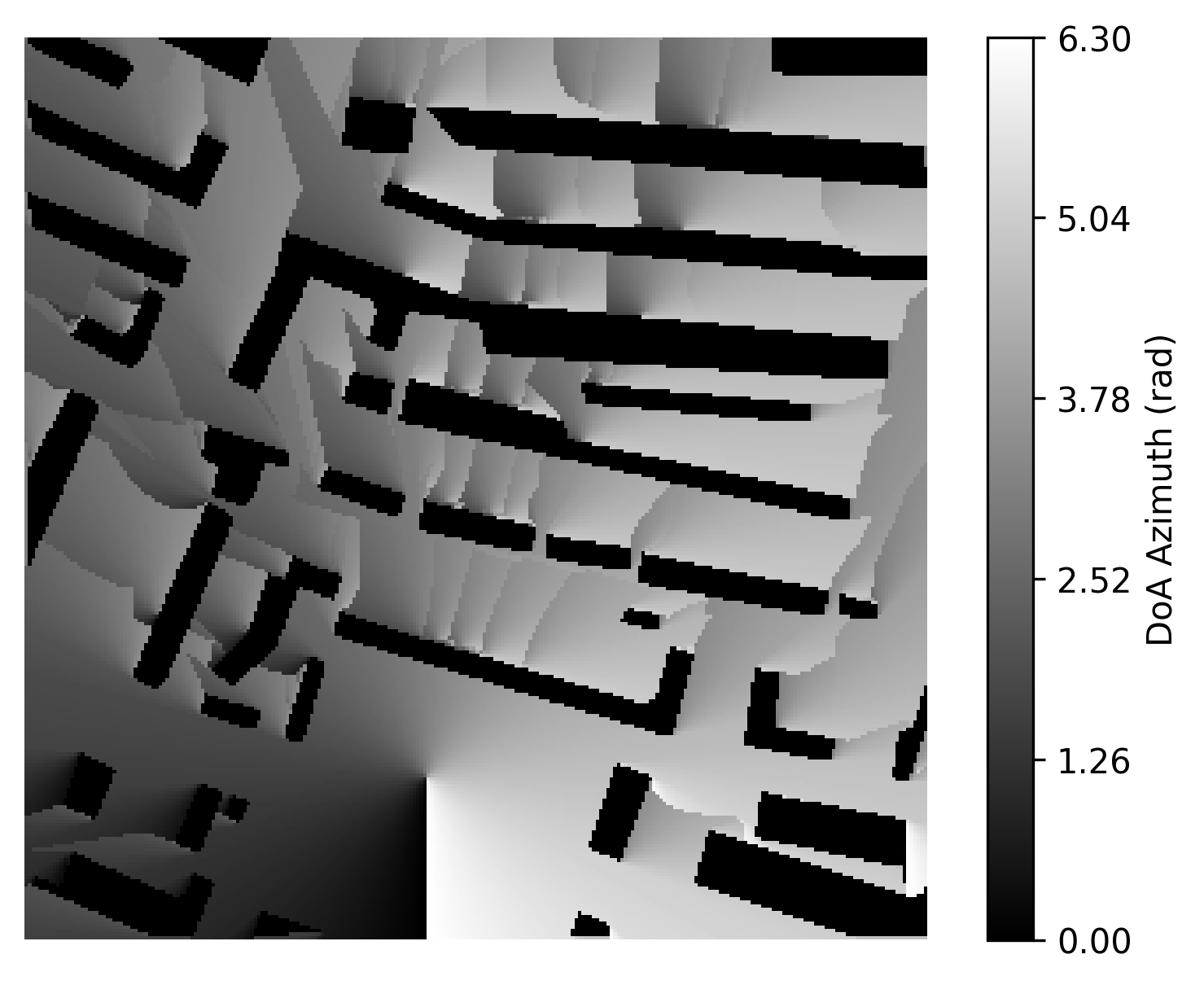}}
  \par\centering
  {\footnotesize \added{DoA Azimuth}}
\end{minipage}

\caption{\added{Measurement results at coordinate \texttt{296\_X46\_Y114} with a receiver height of 3 meters are presented. The figures show spatial maps of multiple radio channel parameters — pathloss, time of arrival (ToA) spread, direction of arrival (DoA) elevation angles, and DoA azimuth angles — all rendered on the same voxel grid. This consistent spatial alignment enables simultaneous visualization and analysis of the interplay among these channel characteristics at the specified location and height. Each figure includes a corresponding colorbar that quantitatively indicates the actual physical values, facilitating precise interpretation of power attenuation, delay dispersion, and angular information. The building regions are assigned a default value of zero, while the remaining spatial voxels reflect valid simulated data.}}

\end{figure*}

\section{Diffusion-Based 3D RM Generation}
\subsection{Problem Formulation of 3D RM Construction}
To support advanced context-aware applications in 6G networks, this work focuses on the construction of 3D multi-modal RMs that capture not only spatial signal propagation but also rich channel characteristics such as angle and ToA information. In contrast to conventional 2D RM representations confined to fixed-height planar pathloss distributions, the proposed framework generalizes the radio mapping task to a volumetric domain, where each spatial point is described by a set of channel parameters across three axes. Specifically, the RM is represented as a 4D tensor \(\mathcal{R} \in \mathbb{R}^{H \times W \times D \times C}\), where \((H, W, D)\) denote the spatial dimensions and \(C\) denotes the number of channel modalities. These modalities include pathloss, DoA, which comprises both azimuth and elevation components, and ToA. This richer representation is essential for enabling key 6G functionalities such as 3D positioning, altitude-sensitive beamforming, and volumetric interference-aware planning. Let the environment be defined as a 3D occupancy grid \(\mathcal{E} \in \{0,1\}^{H \times W \times D}\), where each voxel indicates the presence or absence of physical obstructions (e.g., buildings) with varying shapes and heights. Based on this environmental context, the construction of the 3D RM can be formulated under two practical settings as follows.

\subsubsection{RM Construction with Known Base Station and Environmental Information}   
In this setting, both the environmental geometry \(\mathcal{E}\) and the configuration of the transmitting base station—including its location, elevation, and radiation characteristics—are known. The task is to predict the full RM tensor \(\mathcal{R}\) directly from these inputs. This corresponds to a conditional generative problem, wherein the RM is synthesized based on environmental priors and known source parameters, reflecting plausible channel distributions shaped by the 3D geometry of the scene.

\subsubsection{RM Construction with Sparse Sampling and Environmental Information}  
In this scenario, although the environmental layout \(\mathcal{E}\) is fully known, the transmitter's location and characteristics may be unknown or inaccessible. Instead, a sparse set of signal observations \(\mathcal{S} = \{(x_i, y_i, z_i, \mathbf{r}_i)\}_{i=1}^N\)\ is available, where \(\mathbf{r}_i\) represents partial measurements of channel parameters such as pathloss, DoA, or ToA at specific 3D locations. The objective is to reconstruct the complete RM tensor \(\mathcal{R}\) from these sparse measurements and environmental priors. This problem setting is particularly relevant for spectrum sensing, interference detection, and coverage estimation in scenarios involving unknown or non-cooperative emitters.

By modeling the RM as a structured 3D tensor enriched with angular and temporal descriptors, the proposed framework enables more comprehensive environmental awareness compared to traditional scalar field reconstructions. These two formulations jointly support both transmitter-known and transmitter-agnostic applications, laying the foundation for unified volumetric radio mapping in complex 6G urban environments.

\begin{figure*}[ht]
    \centering
    \includegraphics[width=\textwidth]{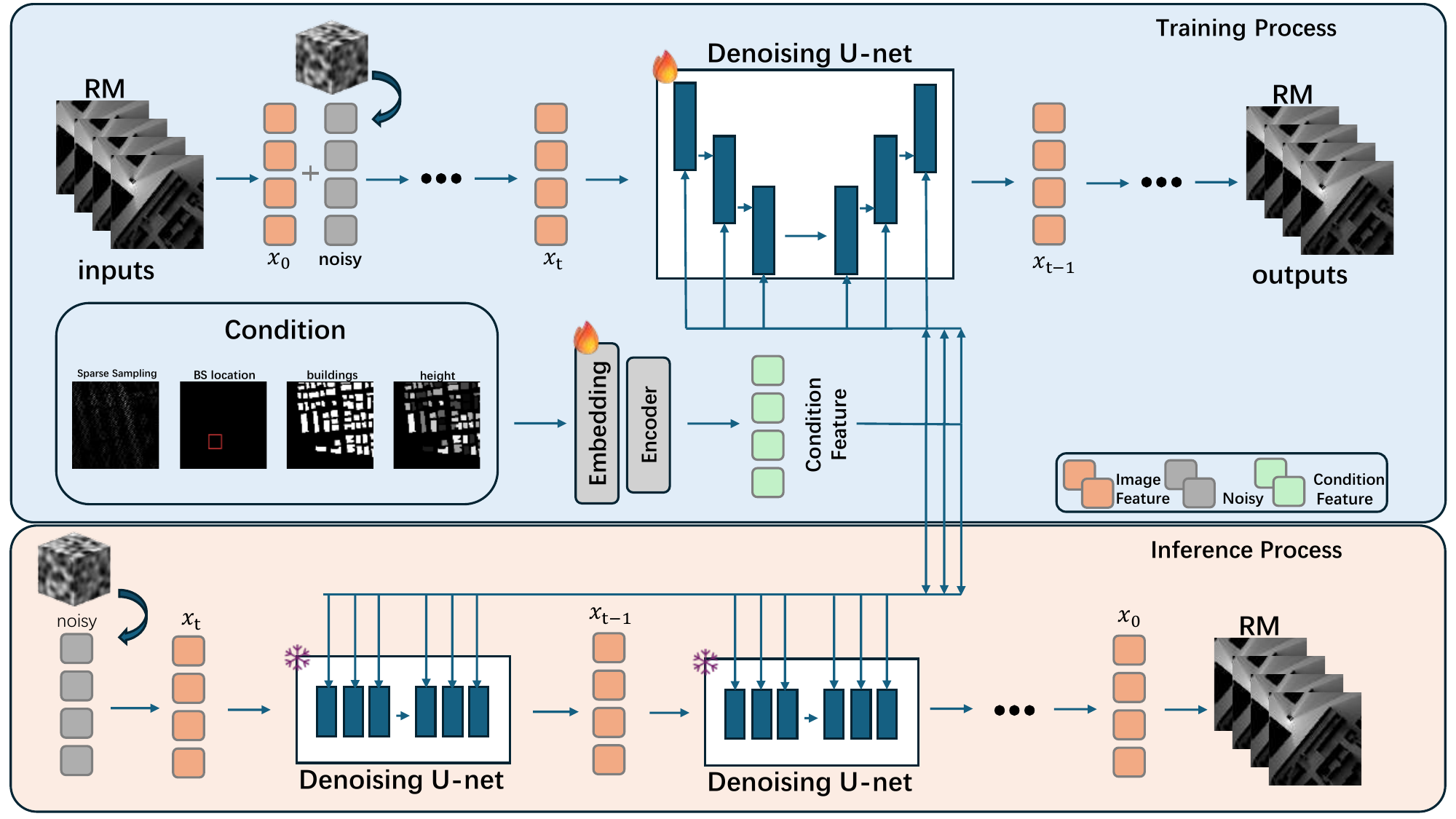}  
    \caption{\added{An overview of the 3D conditional diffusion model framework. During training, the model takes as input the complete 3D radio map and learns to predict the added Gaussian noise using a 3D denoising U-Net. The denoising process is conditioned on auxiliary information, including sparse sampling points, base station (BS) locations, building locations, and building height maps, all of which are provided at each diffusion step. During inference, the model starts from pure noise and progressively generates a full-resolution 3D radio map, guided by the same conditional inputs.}}
\end{figure*}

\subsection{Diffusion Model with 3D Operator}
To enable high-fidelity, multi-modal RM construction in complex 3D environments, we propose RadioDiff-3D, a conditional denoising diffusion probabilistic model (DDPM) designed for volumetric wireless signal generation. In contrast to prior works that estimate pathloss on a fixed-height plane, RadioDiff-3D constructs a 3D tensor-valued RM \(\mathcal{R} \in \mathbb{R}^{H \times W \times D \times C}\), where \((H, W, D)\) represent the spatial dimensions (horizontal, vertical, and height) and \(C\) denotes the number of signal modalities—specifically pathloss, DoA in azimuth and elevation, and ToA. This multi-modal representation is essential for supporting advanced 6G use cases such as 3D beamforming, interference-aware UAV navigation, and real-time volumetric coverage assessment.

\subsubsection{Latent Diffusion Modeling}
Let the input RM tensor \(\mathcal{R}\) be mapped to a latent space via a variational encoder \(\mathcal{E}_{\phi}\), producing a compressed representation \(\bm{z}_0 = \mathcal{E}_{\phi}(\mathcal{R})\). The diffusion process begins by gradually perturbing \(\bm{z}_0\) with Gaussian noise through a forward stochastic process \(\{q(\bm{z}_t|\bm{z}_0)\}_{t=0}^T\), where the transition at each timestep \(t\) is defined as:
\begin{equation}
q(\bm{z}_t | \bm{z}_0) = \mathcal{N}(\bm{z}_t; \sqrt{\bar{\alpha}_t} \bm{z}_0, (1 - \bar{\alpha}_t) \mathbf{I}),
\end{equation}
with \(\bar{\alpha}_t = \prod_{s=1}^{t}(1 - \beta_s)\) being the cumulative noise scaling, and \(\{\beta_t\}_{t=1}^T\) a variance schedule.

The denoising reverse process is learned by training a neural network \(\bm{\epsilon}_\theta(\bm{z}_t, t, \mathcal{C})\) to predict the noise added at each step. The objective is to minimize the expected denoising loss:
\begin{equation}
\mathcal{L}_{\text{simple}} = \mathbb{E}_{\bm{z}_0, \bm{\epsilon}, t} \left[\|\bm{\epsilon}_\theta(\bm{z}_t, t, \mathcal{C}) - \bm{\epsilon}\|^2\right],
\end{equation}
where \(\bm{z}_t = \sqrt{\bar{\alpha}_t} \bm{z}_0 + \sqrt{1 - \bar{\alpha}_t} \bm{\epsilon}\), and \(\mathcal{C}\) denotes the conditioning variables.

\subsubsection{Conditional Generation for Two Application Settings}
RadioDiff-3D supports conditional generation under two distinct application scenarios. In the first scenario, where the environment \(\mathcal{E} \in \{0,1\}^{H \times W \times D}\) (a 3D occupancy grid of buildings) and the base station (BS) configuration \(\bm{r} = (x_{BS}, y_{BS}, z_{BS}, P_t)\) are available, the conditioning vector \(\mathcal{C} = f_{\text{cond}}(\mathcal{E}, \bm{r})\) is passed into each residual block of the denoising U-Net via cross-attention or FiLM-like modulation. This allows the model to learn a distribution \(p_\theta(\bm{z}_0 | \mathcal{E}, \bm{r})\), generating RM tensors aligned with both signal propagation laws and urban topology. In the second scenario, where the BS is unknown or non-cooperative, but a sparse set of signal samples \(\mathcal{S} = \{(\bm{x}_i, \bm{r}_i)\}_{i=1}^N\) is available—where \(\bm{x}_i = (x, y, z)\) and \(\bm{r}_i \in \mathbb{R}^C\) contains pathloss, DoA, and ToA—the model generates \(\mathcal{R}\) conditioned on \((\mathcal{E}, \mathcal{S})\). To enforce coherence between generated and observed regions, we use reconstruction-guided conditional sampling. At inference, the predicted RM is refined as:
\begin{equation}
\tilde{\bm{z}}_t = \bm{z}_t - \lambda_t \nabla_{\bm{z}_t} \left\|\mathcal{R}_\theta(\bm{z}_t) - \mathcal{S}_{\text{interp}}\right\|^2,
\end{equation}
where \(\mathcal{S}_{\text{interp}}\) is the sparsely sampled RM interpolated to match the tensor resolution, and \(\lambda_t\) is a time-dependent guidance weight.

\subsubsection{3D U-Net with Conditioning}
The backbone of RadioDiff-3D is a 3D convolutional U-Net composed of residual blocks and attention layers. The input is a noisy 3D latent tensor \(\bm{z}_t \in \mathbb{R}^{H \times W \times D \times C}\), and the output is either the predicted noise \(\bm{\epsilon}_\theta\) or the denoised latent \(\hat{\bm{z}}_0\). Each residual block integrates conditional embeddings from \(\mathcal{C}\) via cross-attention as follows.
\begin{align}
\text{Attn}(Q, K, V) &= \text{softmax}\left(\frac{QK^T}{\sqrt{d}}\right) V,\\
\quad Q &= W_q F, \\
\quad K,V &= W_k \mathcal{C}, W_v \mathcal{C},
\end{align}
where \(F\) is the current feature map, and \(\mathcal{C}\) is broadcasted or projected to match the spatial resolution. Skip connections bridge encoder and decoder layers, and spectral refinement modules are applied in later stages to capture high-frequency multipath details.

\subsubsection{Autoregressive Height-wise Generation}
To efficiently construct large 3D RM tensors without incurring high memory or compute cost, RadioDiff-3D supports autoregressive vertical generation. Specifically, after generating slices up to height \(z = d - 1\), the model conditions the next slice \(\mathcal{R}_{:,:,d,:}\) on the previous slice \(\mathcal{R}_{:,:,d-1,:}\) via:
\begin{equation}
\mathcal{R}_{:,:,d,:} \sim p_\theta(\mathcal{R}_{:,:,d,:} \mid \mathcal{R}_{:,:,d-1,:}, \mathcal{C}).
\end{equation}
This approach ensures vertical continuity in the generated volume and allows scalable inference across varying building heights and altitudes.


\section{Experimental Process}

This experiment is conducted based on our self-developed 3D diffusion neural network model. To ensure the reproducibility and comparability of results, we adopt a series of standardized experimental settings and rigorously control all aspects of data preprocessing and model training. The following sections provide a detailed description of the experimental procedures and configurations.

\subsection{Preliminary Verification Experiment}
Before conducting the full-scale experiments, we first performed a preliminary verification experiment to validate the feasibility of our proposed method.

In this preliminary study, 10\% of the original training set was randomly selected for training, while 100 samples were drawn from the original testing set for evaluation. The training configuration included a batch size of 2, a learning rate of 1e-4, and 4 frames per input. Automatic Mixed Precision (AMP) was enabled throughout the training process to enhance computational efficiency without sacrificing model performance. Other settings were kept consistent with those used in the full experiments to ensure comparability.

Importantly, the entire preliminary experiment was conducted under a no-sampling setting, meaning that no sampling techniques were applied during the generation process.

\begin{table}[ht]
\caption{Performance Evaluation on Full Sampling Steps}
\centering
\renewcommand{\arraystretch}{1.4}
\resizebox{0.45\textwidth}{!}{%
\begin{tabular}{@{}c|cccc@{}}
\toprule
\textbf{Method}  & \textbf{RMSE} & \textbf{NMSE} & \textbf{SSIM} & \textbf{PSNR}   \\
\midrule
\textbf{No Sampling (1000 steps)} & 0.0653 & 0.0534 & 0.8309 & 24.00  \\
\bottomrule
\end{tabular}
}
\label{tab:inference_time}
\end{table}

Furthermore, we systematically observed the variations in evaluation metrics, including RMSE, NMSE, SSIM, and PSNR. This analysis allowed us to comprehensively assess the reliability and effectiveness of our experimental setup prior to proceeding with the large-scale experiments.

\begin{figure*}[t]
\centering
\setlength\fboxsep{0pt}        
\setlength\fboxrule{0.5pt}     

\begin{minipage}{0.22\textwidth}
\centering
\fbox{\includegraphics[width=\textwidth]{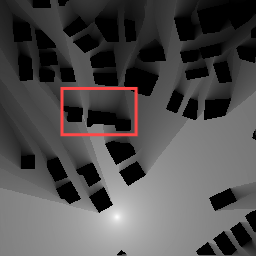}}\\
{\footnotesize \text{Pred\_562\_X39\_Y117\_H1}}
\end{minipage}\hspace{0.1em}
\begin{minipage}{0.22\textwidth}
\centering
\fbox{\includegraphics[width=\textwidth]{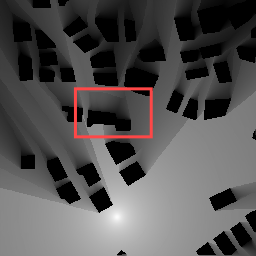}}\\
{\footnotesize \text{Pred\_562\_X39\_Y117\_H2}}
\end{minipage}\hspace{0.1em}
\begin{minipage}{0.22\textwidth}
\centering
\fbox{\includegraphics[width=\textwidth]{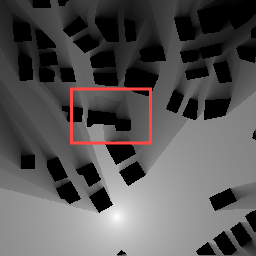}}\\
{\footnotesize \text{Pred\_562\_X39\_Y117\_H3}}
\end{minipage}\hspace{0.1em}
\begin{minipage}{0.22\textwidth}
\centering
\fbox{\includegraphics[width=\textwidth]{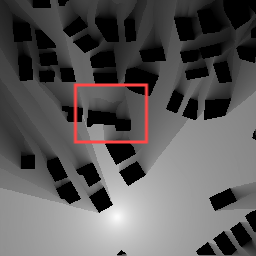}}\\
{\footnotesize \text{Pred\_562\_X39\_Y117\_H4}}
\end{minipage}\\[0.5em]

\begin{minipage}{0.22\textwidth}
\centering
\fbox{\includegraphics[width=\textwidth]{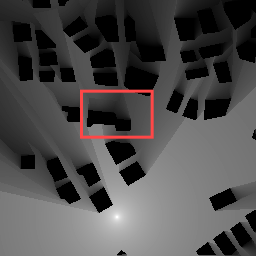}}\\
{\footnotesize \text{GT\_562\_X39\_Y117\_H1}}
\end{minipage}\hspace{0.1em}
\begin{minipage}{0.22\textwidth}
\centering
\fbox{\includegraphics[width=\textwidth]{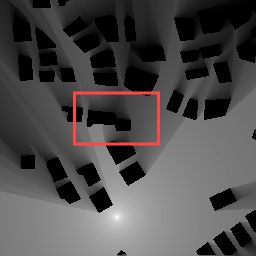}}\\
{\footnotesize \text{GT\_562\_X39\_Y117\_H2}}
\end{minipage}\hspace{0.1em}
\begin{minipage}{0.22\textwidth}
\centering
\fbox{\includegraphics[width=\textwidth]{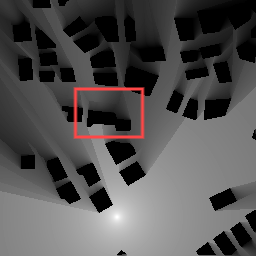}}\\
{\footnotesize \text{GT\_562\_X39\_Y117\_H3}}
\end{minipage}\hspace{0.1em}
\begin{minipage}{0.22\textwidth}
\centering
\fbox{\includegraphics[width=\textwidth]{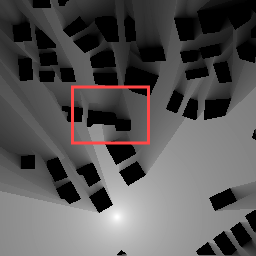}}\\
{\footnotesize \text{GT\_562\_X39\_Y117\_H4}}
\end{minipage}\\[0.5em]

\caption{Comparison between predicted and ground truth frames at heights H=1 to H=4 for the location (562, 39, 117). The top row shows the predicted frames, and the bottom row shows the corresponding ground truth frames.}

\label{fig:res2}
\end{figure*}

To provide a more intuitive understanding of the model performance during this preliminary stage, we visually present the generated video frames and compared frames in Figure~\ref{fig:res2}.

\begin{figure*}[t]
\centering
\setlength\fboxsep{0pt}        
\setlength\fboxrule{0.5pt}     

\begin{minipage}{0.22\textwidth}
\centering
\fbox{\includegraphics[width=\textwidth]{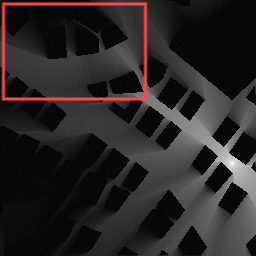}}\\
{\footnotesize \text{Pred\_535\_X93\_Y232\_H1}}
\end{minipage}\hspace{0.1em}
\begin{minipage}{0.22\textwidth}
\centering
\fbox{\includegraphics[width=\textwidth]{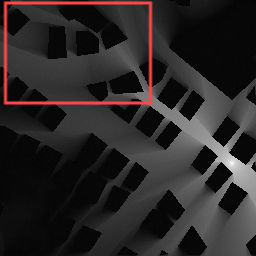}}\\
{\footnotesize \text{Pred\_535\_X93\_Y232\_H2}}
\end{minipage}\hspace{0.1em}
\begin{minipage}{0.22\textwidth}
\centering
\fbox{\includegraphics[width=\textwidth]{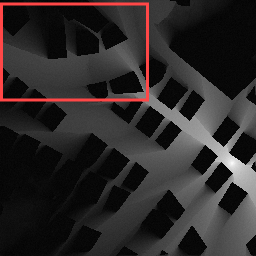}}\\
{\footnotesize \text{Pred\_535\_X93\_Y232\_H3}}
\end{minipage}\hspace{0.1em}
\begin{minipage}{0.22\textwidth}
\centering
\fbox{\includegraphics[width=\textwidth]{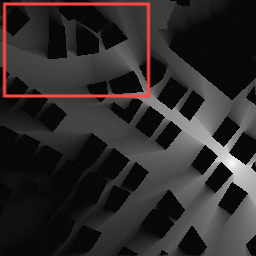}}\\
{\footnotesize \text{Pred\_535\_X93\_Y232\_H4}}
\end{minipage}\\[0.5em]

\begin{minipage}{0.22\textwidth}
\centering
\fbox{\includegraphics[width=\textwidth]{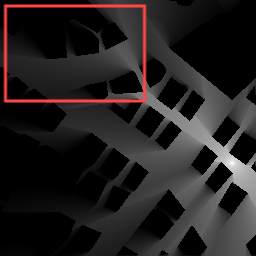}}\\
{\footnotesize \text{GT\_535\_X93\_Y232\_H1}}
\end{minipage}\hspace{0.1em}
\begin{minipage}{0.22\textwidth}
\centering
\fbox{\includegraphics[width=\textwidth]{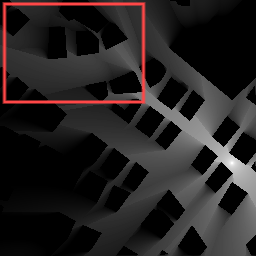}}\\
{\footnotesize \text{GT\_535\_X93\_Y232\_H2}}
\end{minipage}\hspace{0.1em}
\begin{minipage}{0.22\textwidth}
\centering
\fbox{\includegraphics[width=\textwidth]{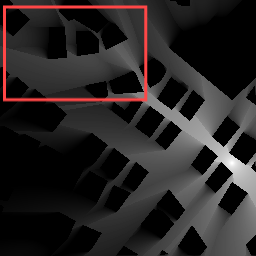}}\\
{\footnotesize \text{GT\_535\_X93\_Y232\_H3}}
\end{minipage}\hspace{0.1em}
\begin{minipage}{0.22\textwidth}
\centering
\fbox{\includegraphics[width=\textwidth]{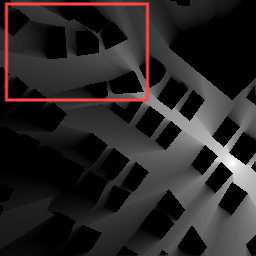}}\\
{\footnotesize \text{GT\_535\_X93\_Y232\_H4}}
\end{minipage}\\[0.5em]

\caption{Comparison between predicted and ground truth frames at heights H=1 to H=4 for the location (535, 93, 232). The top row shows the predicted frames, and the bottom row shows the corresponding ground truth frames.}

\label{fig:res5}
\end{figure*}

\subsection{Dataset and Input Preparation}

To achieve optimal performance in our experiments, we selected specific data from the UrbanRadio3D dataset, focusing on scenarios with heights ranging from 1\,m to 4\,m. From this subset, we extracted the pathloss data as the model output, which was structured as a 4D tensor \( \mathcal{R} \in \mathbb{R}^{H \times W \times 4 \times 1} \).

In the baseline setting without sampling, we used three environmental feature maps as the model input: the building segmentation map, the building height map, and the transmitter location map. These maps were concatenated along the channel dimension to form a 4D input tensor \( \mathcal{R} \in \mathbb{R}^{H \times W \times 4 \times 3} \), serving as the conditional input to the 3D diffusion model.

In experiments where a sampling strategy was employed, we further incorporated a sampling map to provide more detailed spatial information. With this addition, the input tensor was extended to \( \mathcal{R} \in \mathbb{R}^{H \times W \times 4 \times 4} \), where the fourth dimension includes the original three feature maps along with the sampling map. This enhanced configuration enables the model to better capture sparse signal distributions, thereby improving the quality and efficiency of radio map generation.


\added{The original dataset \(\mathcal{D}_{\mathrm{all}}\) consists of \(N\) samples, each corresponding to a unique spatial sampling instance characterized by distinct coordinates and beamforming configurations.}

\added{The dataset is partitioned into two disjoint subsets: the training set \(\mathcal{D}_{\mathrm{train}}\) and the testing set \(\mathcal{D}_{\mathrm{test}}\), such that}
\begin{equation}
\added{
\mathcal{D}_{\mathrm{all}} = \mathcal{D}_{\mathrm{train}} \cup \mathcal{D}_{\mathrm{test}}, \quad
\mathcal{D}_{\mathrm{train}} \cap \mathcal{D}_{\mathrm{test}} = \emptyset,
}
\end{equation}

\added{with}
\begin{equation}
\added{
|\mathcal{D}_{\mathrm{train}}| = \left\lfloor 0.9 \times N \right\rfloor, \quad
|\mathcal{D}_{\mathrm{test}}| = N - |\mathcal{D}_{\mathrm{train}}|.
}
\end{equation}

\added{The splitting is performed randomly at the file level by shuffling the sample indices and assigning the first 90\% to the training set and the remaining 10\% to the testing set.}

\added{Due to the uniqueness of each sample's spatial coordinate and beamforming setup, the intersection between \(\mathcal{D}_{\mathrm{train}}\) and \(\mathcal{D}_{\mathrm{test}}\) is empty, ensuring no spatial or semantic overlap exists between the two subsets.}

\added{This strict separation guarantees statistical independence and eliminates the possibility of information leakage from training to testing phases.}






\subsection{Sampling Strategies}
To study the influence of the sampling strategies on the RM construction performance, we conduct experiments of RM construction with sampling information. These strategies aim to sparsify the input GIF tensor by selecting representative pixels. \added{During model training, we applied both uniform and random sampling strategies to sparsely select data points from the constructed 3D radio maps. A sampling rate of 10\% was adopted, implying that for a 3D tensor consisting of 256×256 pixels per slice, approximately 6,553 voxels were used for training at each height level. This sampling density was selected to strike a balance between computational tractability and model generalization capability.}

\subsubsection{Uniform Sampling}
For uniform sampling, a fixed number of pixels are selected uniformly from each frame to form a sparse GIF tensor. Given the tensor \( \mathbf{R} \in \mathbb{R}^{H \times W \times D \times C} \), where:
\begin{itemize}
  \item \textit{\( H \) and \( W \)} are the height and width of each frame;
  \item \textit{\( D \)} is the number of frames;
  \item \textit{\( C \)} is the number of channels.
\end{itemize}
Let \( \mathcal{I} \subseteq \{1, \dots, H \times W \} \) represent the set of indices for the pixels uniformly sampled from each frame. The uniform sampling operation can be mathematically expressed as:

\begin{equation}
\mathbf{S}_{\text{uniform}}(h, w, d, c) = 
\begin{cases} 
\mathbf{R}(h, w, d, c) & \text{if } (h, w) \in \mathcal{I}_{\text{uniform}}, \\
0 & \text{otherwise}.
\end{cases}
\end{equation}
where \( \mathbf{S}_{\text{uniform}} \) denotes the resulting sparse tensor, with the sampled pixels retained as they are from the original tensor, while non-sampled pixels are set to a value of 0.

\subsubsection{Random Sampling}
Random sampling involves selecting a random subset of pixels from each frame, generating a sparse gif tensor. This can be represented as follows:

\begin{equation}
\mathbf{S}_{\text{random}}(h, w, d, c) = 
\begin{cases} 
\mathbf{R}(h, w, d, c) & \text{if } (h, w) \in \mathcal{I}_{\text{random}}, \\
0 & \text{otherwise}.
\end{cases}
\end{equation}

Where \( \mathcal{I}_{\text{random}} \) represents the randomly selected pixel indices from the frame.









\subsection{Experimental Parameter Configuration}

The training process was configured with several key hyperparameters to optimize model performance. Specifically, a batch size of 2 was used, with a learning rate set to 5e-5, and an exponential moving average (EMA) decay rate of 0.995. To enhance computational efficiency, automatic mixed precision (AMP) was enabled, and the EMA was updated every 10 iterations. The L1 loss function, known for its effectiveness in regression tasks, was employed. The model was trained on an NVIDIA 4090 device. These hyperparameters were carefully chosen to strike a balance between training speed and stability, while ensuring that the model’s convergence and generalization capabilities were optimized through the use of AMP and EMA.

\subsection{Performance Metrics}
To comprehensively evaluate the reconstruction quality of the generated RMs, we conduct both qualitative and quantitative analyses. Figure~\ref{fig:res5} presents the visual comparison between the predicted and ground truth RMs across various altitudes, revealing distinct differences in spatial structure and signal distribution. These qualitative observations are quantitatively supported by several widely adopted image quality metrics, including root mean squared error (RMSE), normalized mean squared error (NMSE), structural similarity index metric (SSIM), and peak signal-to-noise ratio (PSNR), as summarized in Table~\ref{tab:inference_time}.

\begin{table}[ht]
\caption{Evaluation Metrics for DDIM Sampling (200 Steps) Without Additional Sampling Strategy}
\centering
\renewcommand{\arraystretch}{1.4}
\begin{tabular}{@{}c|cccc@{}}
\toprule
\textbf{Sampling Method} & \textbf{RMSE} & \textbf{NMSE} & \textbf{SSIM} & \textbf{PSNR} \\
\midrule
\textbf{No Sampling} & 0.1325 & 0.3472 & 0.6453 & 19.57 \\
\textbf{Sampling (rate 10\%)} & 0.0481 & 0.0550 & 0.8187 & 29.23 \\
\bottomrule
\end{tabular}
\label{tab:inference_time-1}
\end{table}

\subsubsection{MSE}
MSE is calculated by averaging the squared differences between the pixel intensities of the original and final images, as follows:
\begin{equation}MSE=\frac1{NM}\Sigma_{m=0}^{M-1}\sum_{n=0}^{N-1}e(m,n)^2,\end{equation}
where \( e(m, n) \) is the error difference between the ground truth and the predicted image, and \( M \) and \( N \) are the height and width of the image, respectively. The NMSE is a scaled version of MSE used to evaluate predictive accuracy. When constructing the RM, the RMSE is simply the square root of MSE, which is defined as follows:

\begin{figure*}[t]
\centering
\setlength\fboxsep{0pt}        
\setlength\fboxrule{0.5pt}     

\begin{minipage}{0.16\textwidth}
\centering
\fbox{\includegraphics[width=\textwidth]{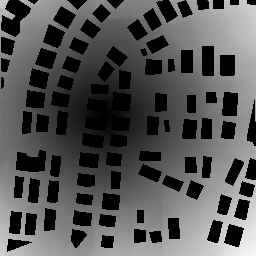}}\\
{\footnotesize \text{Pred\_2\_X139\_Y104\_H1}}
\end{minipage}\hspace{0.1em}
\begin{minipage}{0.16\textwidth}
\centering
\fbox{\includegraphics[width=\textwidth]{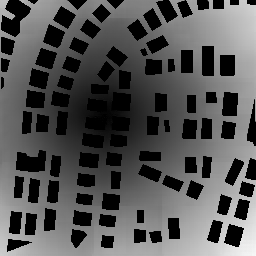}}\\
{\footnotesize \text{Pred\_2\_X139\_Y104\_H3}}
\end{minipage}\hspace{1.5em}
\begin{minipage}{0.16\textwidth}
\centering
\fbox{\includegraphics[width=\textwidth]{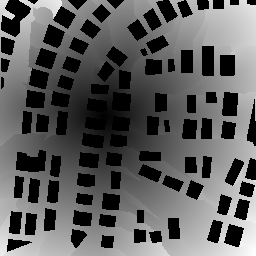}}\\
{\footnotesize \text{GT\_2\_X139\_Y104\_H1}}
\end{minipage}\hspace{0.1em}
\begin{minipage}{0.16\textwidth}
\centering
\fbox{\includegraphics[width=\textwidth]{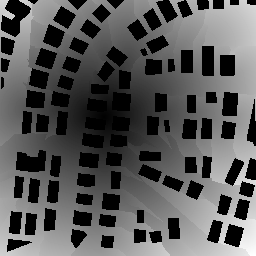}}\\
{\footnotesize \text{GT\_2\_X139\_Y104\_H3}}
\end{minipage}\\[0.5em]

\begin{minipage}{0.16\textwidth}
\centering
\fbox{\includegraphics[width=\textwidth]{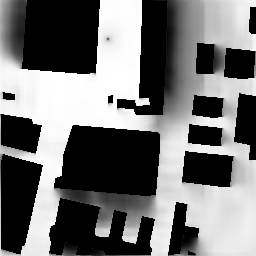}}\\
{\footnotesize \text{Pred\_105\_X217\_Y108\_H1}}
\end{minipage}\hspace{0.1em}
\begin{minipage}{0.16\textwidth}
\centering
\fbox{\includegraphics[width=\textwidth]{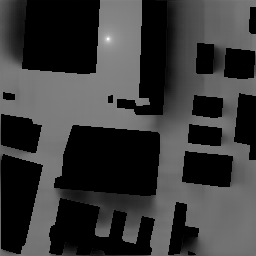}}\\
{\footnotesize \text{Pred\_105\_X217\_Y108\_H3}}
\end{minipage}\hspace{1.5em}
\begin{minipage}{0.16\textwidth}
\centering
\fbox{\includegraphics[width=\textwidth]{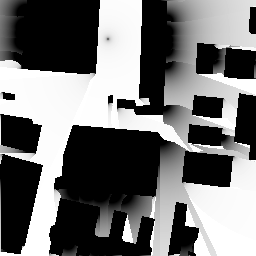}}\\
{\footnotesize \text{GT\_105\_X217\_Y108\_H1}}
\end{minipage}\hspace{0.1em}
\begin{minipage}{0.16\textwidth}
\centering
\fbox{\includegraphics[width=\textwidth]{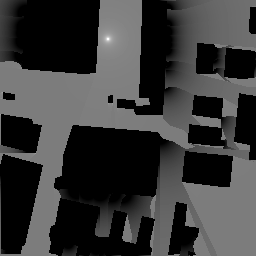}}\\
{\footnotesize \text{GT\_105\_X217\_Y108\_H3}}
\end{minipage}\\[0.5em]

\begin{minipage}{0.16\textwidth}
\centering
\fbox{\includegraphics[width=\textwidth]{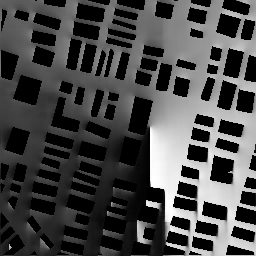}}\\
{\footnotesize \text{Pred\_48\_X129\_Y150\_H1}}
\end{minipage}\hspace{0.1em}
\begin{minipage}{0.16\textwidth}
\centering
\fbox{\includegraphics[width=\textwidth]{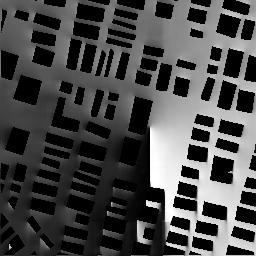}}\\
{\footnotesize \text{Pred\_48\_X129\_Y150\_H3}}
\end{minipage}\hspace{1.5em}
\begin{minipage}{0.16\textwidth}
\centering
\fbox{\includegraphics[width=\textwidth]{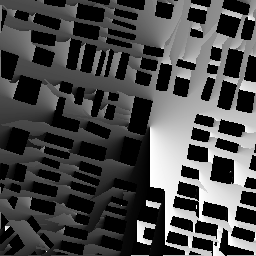}}\\
{\footnotesize \text{GT\_48\_X129\_Y150\_H1}}
\end{minipage}\hspace{0.1em}
\begin{minipage}{0.16\textwidth}
\centering
\fbox{\includegraphics[width=\textwidth]{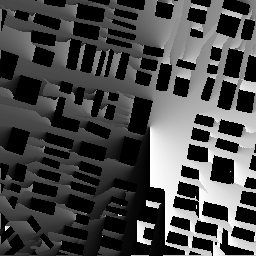}}\\
{\footnotesize \text{GT\_48\_X129\_Y150\_H3}}
\end{minipage}\\[0.5em]

\caption{Visual comparison of 3D-UNet predictions and ground truth for Delay, DoA\_Ele, and DoA\_Azi at 1\,m (H1) and 3\,m (H3) heights. Each row shows predictions and corresponding ground truth side by side for qualitative evaluation.}

\label{fig:res6}
\end{figure*}

\begin{equation}NMSE=\frac{\Sigma_{m=1}^M\Sigma_{n=1}^N(I_b(m,n)-I(m,n))^2}{\Sigma_{m=1}^M\sum_{n=1}^NI^2(m,n)},\end{equation}
\begin{equation}\mathrm{RMSE}=\sqrt{MSE}\end{equation}

\subsubsection{SSIM}
SSIM is a quality assessment metric inspired by the human visual system. Since SSIM primarily focuses on measuring texture differences and there are many high-frequency details in RM (Reconstructed Map), SSIM is suitable for evaluating the quality of the generated results. Additionally, we believe more attention should be paid to the brightness of signal radiation, the contrast between signal radiation and surrounding areas, and the accuracy of geographical maps in RM reconstruction. This aligns with the SSIM metric, which evaluates three key components: brightness, contrast, and structural information, calculated as follows:
\begin{equation}
l(x, y) = \frac{2\mu_X(x, y) \mu_Y(x, y) + C_1}{\mu_X^2(x, y) + \mu_Y^2(x, y) + C_1} 
\end{equation}

\begin{equation}
c(x, y) = \frac{2\sigma_X(x, y)\sigma_Y(x, y) + C_2}{\sigma_X^2(x, y) + \sigma_Y^2(x, y) + C_2} 
\end{equation}

\begin{equation}
s(x, y) = \frac{\sigma_{XY}(x, y) + C_3}{\sigma_X(x, y) \sigma_Y(x, y) + C_3} 
\end{equation}

Where \(x\) and \(y\) correspond to two different input images, and \(\mu_x\), \(\sigma_x^2\), \(\sigma_{xy}\) represent the mean, variance of \(x\), and the covariance between \(x\) and \(y\), respectively. Additionally, \(C_1\), \(C_2\), and \(C_3\) are constants defined as:

\[
C_1 = (K_1 L)^2, \quad C_2 = (K_2 L)^2, \quad C_3 = \frac{C_2}{2}
\]

Where \(L\) represents the dynamic range of the data. Based on these parameters, the structural similarity can be computed as follows:

\begin{equation}
SSIM(x, y) = \frac{(2\mu_x \mu_y + C_1)(\sigma_{xy} + C_2)}{(\mu_x^2 + \mu_y^2 + C_1)(\sigma_x^2 + \sigma_y^2 + C_2)} 
\end{equation}

\subsubsection{PSNR}
PSNR is defined as the ratio between the maximum possible power of a signal and the power of the noise that interferes with its representation accuracy. PSNR is typically expressed in decibels (dB) and provides an approximate measure of the perceived quality of a reconstruction. In image evaluation, a higher PSNR generally indicates better image quality. For RMs, an accurate edge signal is critical; therefore, PSNR is used not only to assess the overall image quality but also to evaluate the quality of edge details in the generated RMs. PSNR can be calculated as $PSNR=10\log_{10}\left(\frac{r^2}{MSE}\right)$.

\subsection{Inference Time}

To evaluate the computational efficiency of the model during inference, we measured the time required to generate a complete RM under different DDIM sampling steps. As shown in Table~\ref{tab:inference_time1}, the inference time increases approximately linearly with the number of sampling steps. Specifically, using 20 steps yields a rapid inference time of 2.43 seconds, while increasing to 200 steps results in a significantly longer time of 24.35 seconds. This highlights a clear trade-off between reconstruction quality and computational cost, which is crucial for real-time or resource-constrained deployment scenarios.

\begin{table}[ht]
\caption{Inference Time for DDIM Sampling at Various Step Counts}
\centering
\renewcommand{\arraystretch}{1.4}
\resizebox{0.33\textwidth}{!}{%
\begin{tabular}{@{}c|c|c@{}}
\toprule
\textbf{Method} & \textbf{Steps} & \textbf{Inference Time (s)} \\
\midrule
\multirow{5}{*}{DDIM Sampling} 
& 20  & 2.4300  \\
& 30  & 3.6574  \\
& 50  & 6.1171  \\
& 100 & 12.5732 \\
& 200 & 24.3526 \\
\bottomrule
\end{tabular}
}
\label{tab:inference_time1}
\end{table}

\subsection{3D-UNet Architecture Overview}
\added{3D-UNet\cite{iek20163DUL} is an extension of the traditional UNet architecture baseline model due to its proven effectiveness in radio map construction, as demonstrated by prior works such as RadioUNet \cite{levie2021radiounet}, RME-GAN \cite{zhang2023rme}, and RadioDiff \cite{wang2024radiodiff}, all of which employ CNN-based architectures. The use of 3D convolutional operators further enables the model to efficiently capture spatial correlations in volumetric wireless environments}. While the standard 2D UNet is widely used for image segmentation tasks, 3D-UNet extends its capabilities to three-dimensional data by replacing 2D convolutional and pooling operations with their 3D counterparts. This design allows the model to capture spatial context across depth, height, and width, making it particularly suitable for tasks involving volumetric or temporal sequences, such as medical imaging or spatiotemporal modeling.

The architecture retains the encoder-decoder structure of the original UNet, where the encoder progressively reduces spatial resolution to capture high-level features, and the decoder symmetrically reconstructs the output while integrating low-level spatial information via skip connections. This combination enables precise localization and accurate representation of complex spatial relationships within the data.

Key advantages of 3D-UNet include:
\begin{itemize}
    \item Enhanced spatial feature learning by considering the depth dimension.
    \item Effective modeling of volumetric correlations and contextual dependencies.
    \item Strong performance on tasks requiring dense prediction over 3D space.
\end{itemize}

\subsection{Application of 3D-UNet in Our Experiment}
In our experiment, we employ a 3D-UNet model to learn the mapping from environmental features to wireless propagation characteristics. The network input comprises three spatial feature maps: a Building Segmentation Map, a Building Height Map, and a Transmitter Location Map. These maps are stacked along the channel dimension to form a 4D tensor \( \mathcal{R} \in \mathbb{R}^{H \times W \times 4 \times 3} \), where \( H \) and \( W \) denote the spatial resolution, the depth dimension (4) corresponds to discrete altitude levels (1m, 2m, 3m, and 4m), and the channel dimension (3) represents the input features. The model is trained to predict three wireless channel characteristics: the ToA Map, the DoA in azimuth (DoA\_Azi), and the DoA in elevation (DoA\_Ele). Each target is represented as a separate 4D tensor of shape \( \mathcal{R} \in \mathbb{R}^{H \times W \times 4 \times 1} \), sharing the same spatial and altitude resolution as the input. 

\begin{table}[ht]
\caption{Evaluation Metrics on Each Output}
\centering
\renewcommand{\arraystretch}{1.4}
\begin{tabular}{@{}c|c|cccc@{}}
\toprule
\textbf{Methods} & \textbf{Output} & \textbf{RMSE} & \textbf{NMSE} & \textbf{SSIM} & \textbf{PSNR} \\
\midrule
\multirow{3}{*}{No sampling} & \textbf{ToA}     & 0.0287 & 0.0046 & 0.9740 & 31.25 \\
                             & \textbf{DoA\_Azi}  & 0.0921 & 0.0496 & 0.8603 & 21.85 \\
                             & \textbf{DoA\_Ele}  & 0.0744 & 0.0227 & 0.9156 & 24.27 \\
\midrule 
\multirow{3}{*}{\shortstack{Random sampling \\ (sampling rate 10\%)}} & \textbf{ToA}     & 0.0147 & 0.0012 & 0.9881 & 36.98 \\
                             & \textbf{DoA\_Azi}  & 0.0362 & 0.0070 & 0.9582 & 29.55 \\
                             & \textbf{DoA\_Ele}  & 0.0317 & 0.0039 & 0.9669 & 31.45 \\
\bottomrule

\multirow{3}{*}{\shortstack{Uniform sampling \\ (sampling rate 10\%)}} & \textbf{ToA}     & 0.0140 & 0.0010 & 0.9849 & 37.39 \\
                             & \textbf{DoA\_Azi}  & 0.0299 & 0.0047 & 0.9677 & 31.18 \\
                             & \textbf{DoA\_Ele}  & 0.0227 & 0.0021 & 0.9783 & 34.29 \\
                             
\bottomrule
\end{tabular}
\label{tab:output_metrics}
\end{table}

\subsection{3D-UNet Performance Metrics}

We evaluate the performance of our 3D-UNet-based model using four standard metrics as RMSE, NMSE, SSIM, and PSNR. These metrics capture different aspects of prediction quality, including accuracy, structural fidelity, and perceptual quality. Evaluation is conducted on 1000 test samples that are disjoint from the training set. The results for the predicted ToA, DoA\_Azi, and DoA\_Ele are summarized in Table~\ref{tab:output_metrics}.

A visual comparison is shown in Fig.~\ref{fig:res6}, which aligns with the quantitative metrics.


\section{Conclusion}
In this paper, we have introduced UrbanRadio3D, a high-resolution 3D×3D radio map dataset that captures spatial, angular, and temporal propagation characteristics in realistic urban environments. To explore the feasibility of learning-based volumetric RM construction, we have established benchmark models, including a conditional diffusion model, RadioDiff-3D, and a conventional 3D convolutional neural network, 3D-UNet, and have evaluated them across pathloss, DoA, and ToA modalities under both transmitter-aware and transmitter-agnostic settings. These results have demonstrated the viability of data-driven 3D RM generation and provide strong baselines for future research. The proposed dataset and framework offer a foundational toolset for advancing environment-aware wireless communication and intelligent decision-making in 6G networks. Future work will focus on extending the dataset to multi-band scenarios and integrating physical priors into generative modeling. \added{Moreover, the integration of volumetric transformer architectures holds strong potential for enhancing long-range spatial modeling capabilities, especially in dynamic or large-scale urban environments where non-local dependencies are prominent.}

\bibliography{ref}
\bibliographystyle{IEEEtran}

\begin{IEEEbiography}[{\includegraphics[width=1in,height=1.25in,clip,keepaspectratio]{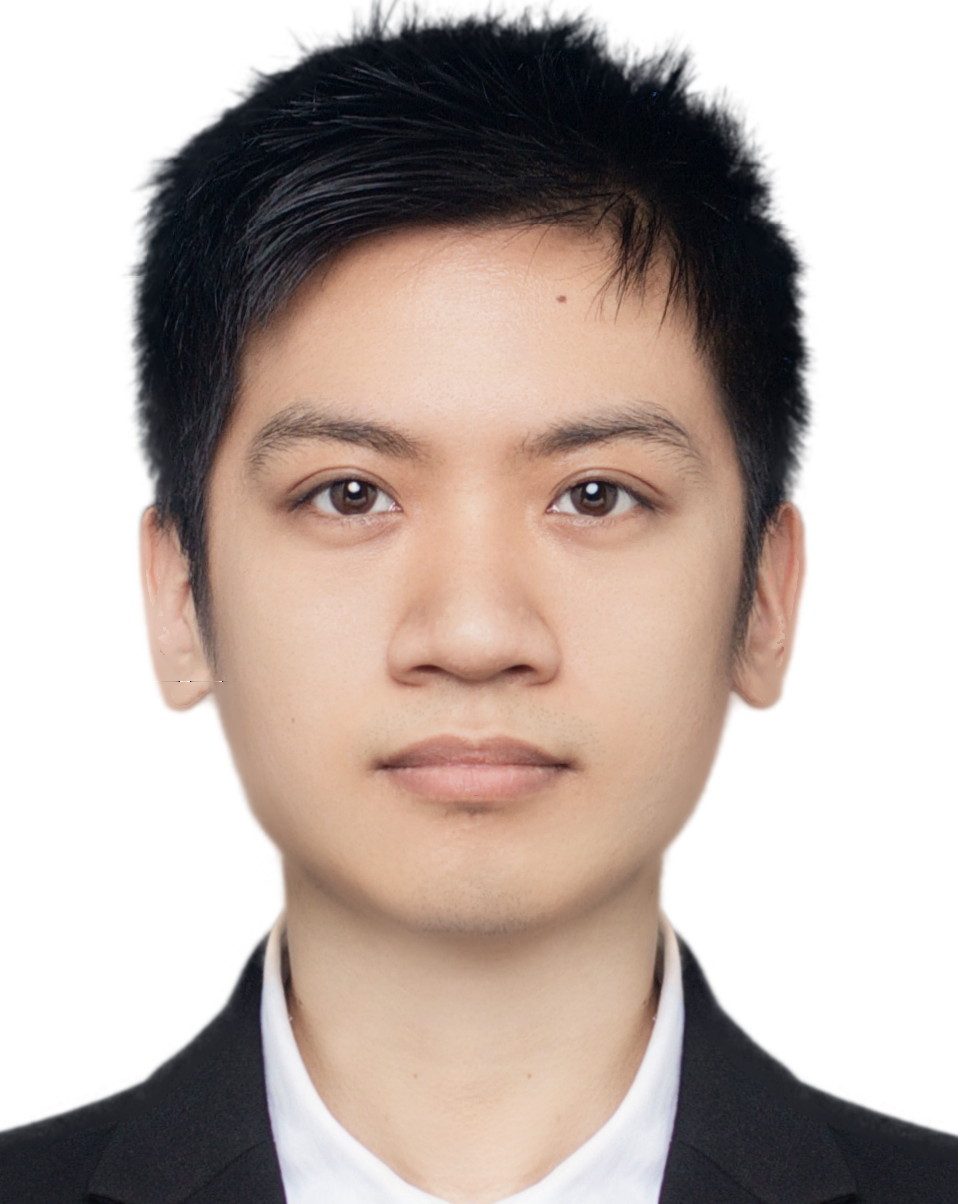}}]{Xiucheng Wang}
is currently pursuing a Ph.D degree from Xidian University. His research areas of
interest are radio maps, generative artificial intelligence, and channel estimation.
\end{IEEEbiography}

\begin{IEEEbiography}[{\includegraphics[width=1in,height=1.25in,clip,keepaspectratio]{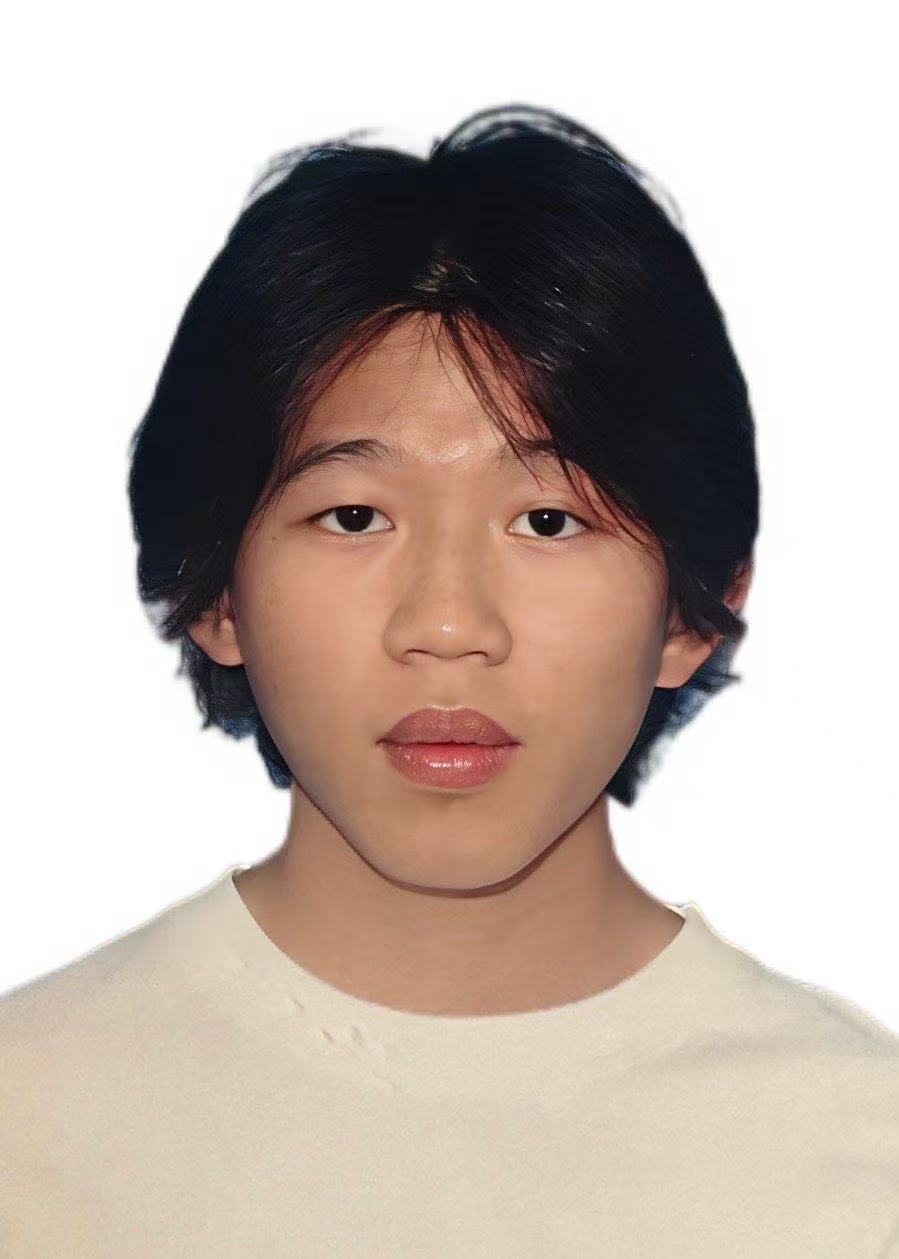}}]{Qiming Zhang}
is currently pursuing an undergraduate degree at Xidian University, Xi’an, China. His research focuses on diffusion models and radio maps.
\end{IEEEbiography}

\begin{IEEEbiography}[{\includegraphics[width=1in,height=1.25in,clip,keepaspectratio]{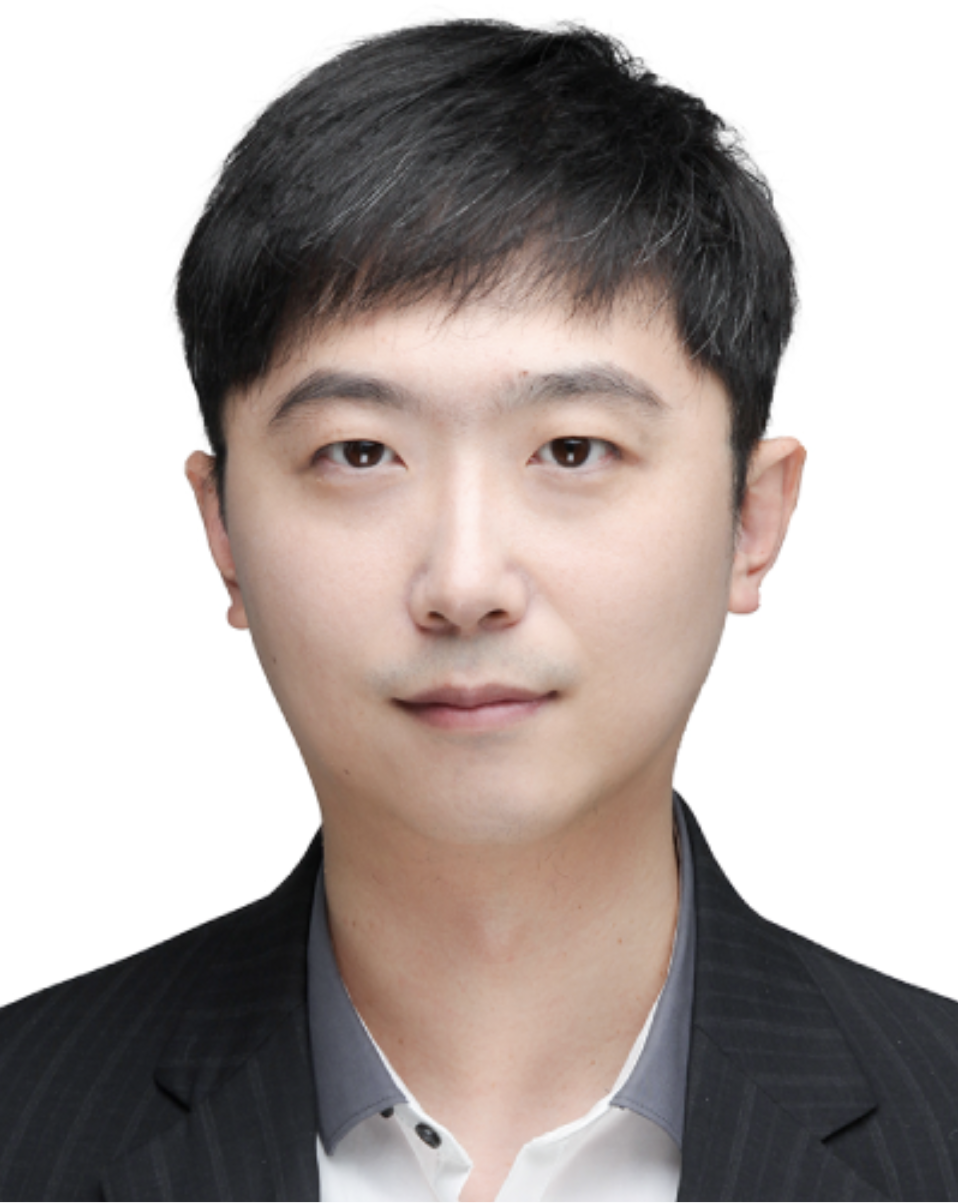}}]{Nan Cheng}
received the B.E. and M.S. degrees from the Department of Electronics and Information Engineering, Tongji University, Shanghai, China, in 2009 and 2012, respectively, and the Ph.D. degree from the Department of Electrical and Computer Engineering, University of Waterloo, Waterloo, ON, Canada, in 2016. From 2017 to 2019, he was a Postdoctoral Fellow with the Department of Electrical and Computer Engineering, University of Toronto, Toronto, ON, Canada. He is currently a Professor with the State Key Laboratory of ISN and the School of Telecommunications Engineering, Xidian University, Xi’an, Shaanxi, China. He has authored or co-authored more than 90 journal papers in IEEE Transactions and other top journals. His research interests include B5G/6G, AI-driven future networks, and space-air–ground-integrated networks. Prof. Cheng is an Associate Editor of the IEEE Transactions on Vehicular Technology, IEEE Open Journal of the Communications Society, and Peer-to-Peer Networking and Applications. He is/was the guest editor of several journals.
\end{IEEEbiography}

\begin{IEEEbiography}[{\includegraphics[width=1in,height=1.25in,clip,keepaspectratio]{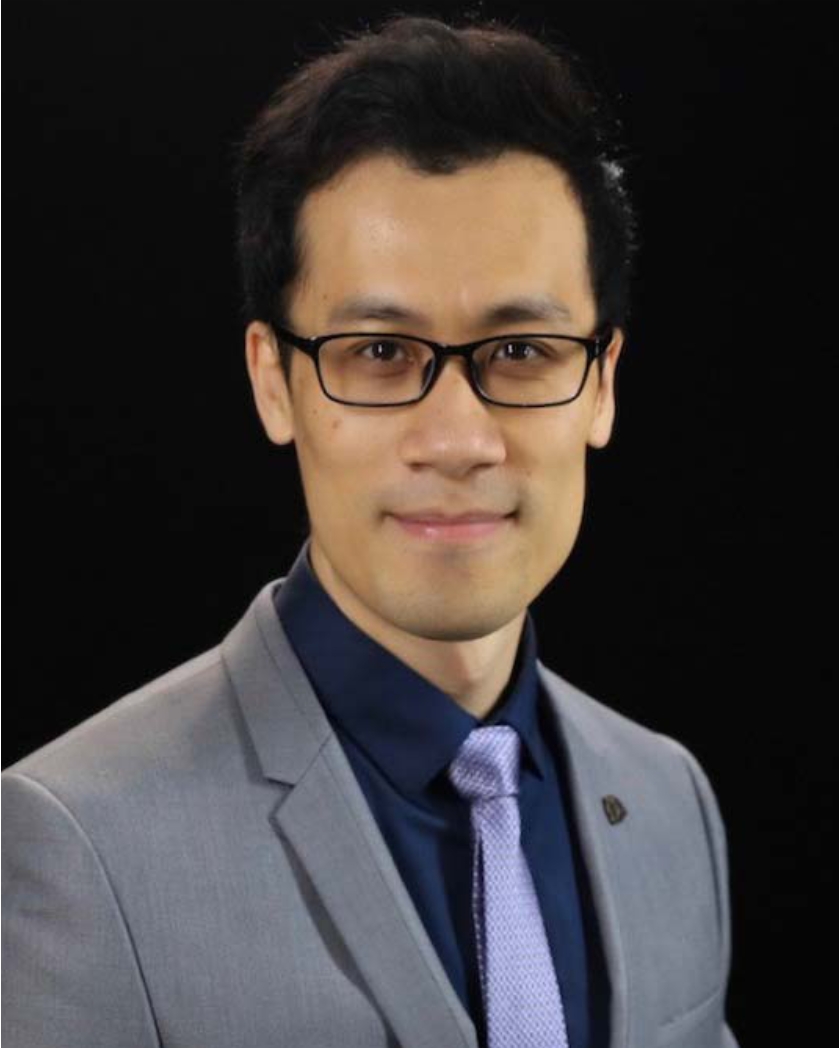}}]{Junting Chen}
received the B.Sc. degree in electronic engineering from Nanjing University, Nanjing, China, in 2009, and the Ph.D. degree in electronic and computer engineering from The Hong Kong University of Science and Technology (HKUST), Hong Kong, SAR, China, in 2015. From 2014 to 2015, he was a Visiting Student with the Wireless Information and Network Sciences Laboratory, MIT, Cambridge, MA, USA. He is currently an Assistant Professor with the School of Science and Engineering and the Future Network of Intelligence Institute (FNii), The Chinese University of Hong Kong (CUHK), Shenzhen, Guangdong, China. Prior to joining CUHK, he was a Post-Doctoral Research Associate with the Communication Systems Department, Eurecom, Sophia Antipolis, France, from 2015 to 2016, and the Ming Hsieh Department of Electrical Engineering, University of Southern California (USC), Los Angeles, CA, USA, from 2016 to 2018. His research interests include channel estimation, MIMO beamforming, machine learning, optimization for wireless communications and localization, radio map sensing, construction, and application for wireless communications. He was a recipient of the HKTIIT Post-Graduate Excellence Scholarships in 2012. He was nominated as the Exemplary Reviewer of IEEE Wireless Communications Letters in 2018. His article received the Charles Kao Best Paper Award from WOCC 2022.
\end{IEEEbiography}

\begin{IEEEbiography}[{\includegraphics[width=1in,height=1.25in,clip,keepaspectratio]{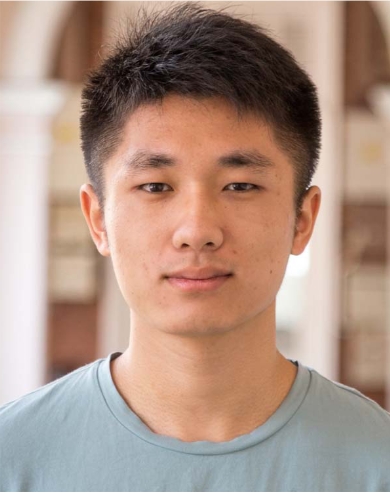}}]{Zezhong Zhang}
is currently a Research Assistant Professor with the School of Science and Engineering (SSE), the Shenzhen Future Network of Intelligence Institute (FNii-Shenzhen), and the Guangdong Provincial Key Laboratory of Future Networks of Intelligence, The Chinese University of Hong Kong, Shenzhen, China. His research interests are in the areas of edge learning, radio map estimation, integrated sensing and communication and B5G technologies.
\end{IEEEbiography}

\begin{IEEEbiography}[{\includegraphics[width=1in,height=1.25in,clip,keepaspectratio]{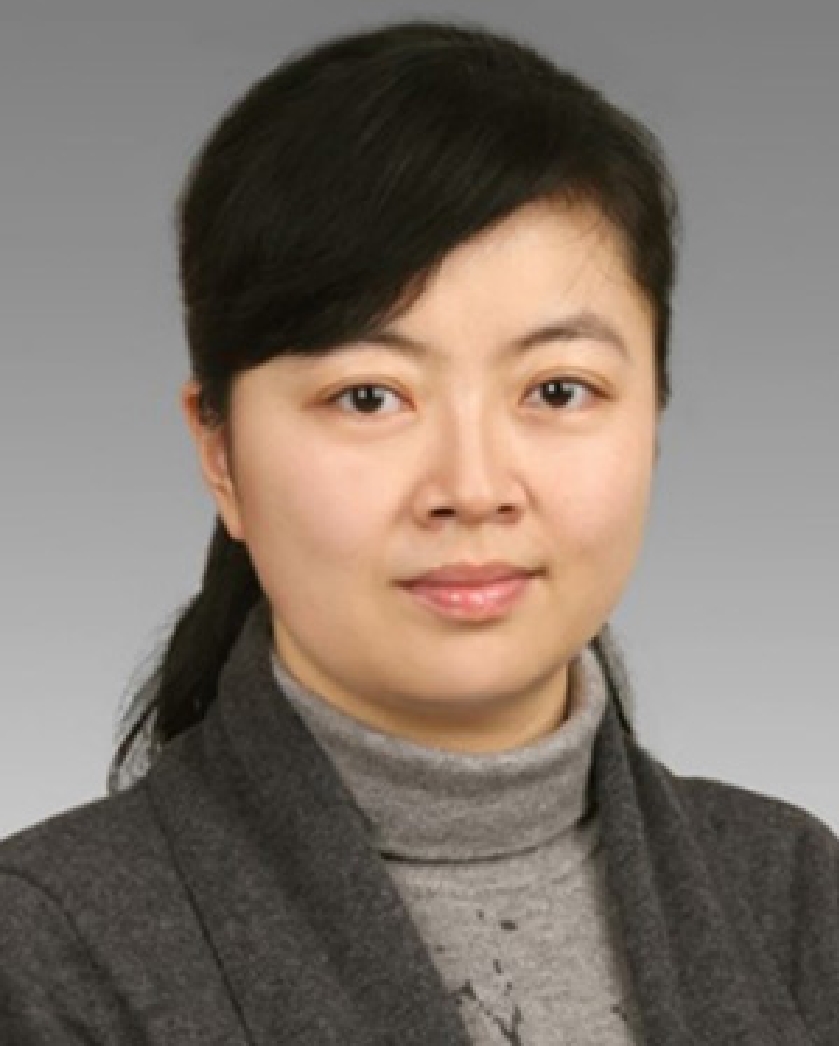}}]{Zan Li}
received the B.S. degree in communications engineering and the M.S. and Ph.D. degrees in communication and information systems from Xidian University, Xi’an, China, in 1998, 2001, and 2006, respectively. She is currently a Professor with the State Key Laboratory of Integrated Services Networks, School of Telecommunications Engineering, Xidian University. Her research interests include topics on wireless communications and signal processing, such as covert communication, spectrum sensing, and cooperative communications.,Prof. Li was awarded the National Science Fund for Distinguished Young Scholars. She serves as an Associate Editor for the IEEE Transactions on Cognitive Communications and Networking and China Communications. She is a Fellow of the Institution of Engineering and Technology, the China Institute of Electronics, and the China Institute of Communications.
\end{IEEEbiography}

\begin{IEEEbiography}[{\includegraphics[width=1in,height=1.25in,clip,keepaspectratio]{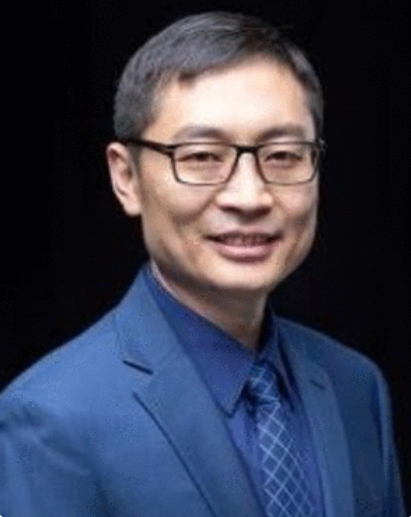}}]{Shuguang Cui}
received the Ph.D. degree in electrical engineering from Stanford University, Stanford, CA, USA, in 2005. He was an Assistant Professor, an Associate Professor, a Full Professor, and the Chair Professor of Electrical and Computer Engineering with the University of Arizona, Texas A\&M University, UC Davis, and CUHK-Shenzhen, Shenzhen, China, respectively. He was the Executive Dean of the School of Science and Engineering, CUHK-Shenzhen, and the Director of the Future Network of Intelligence Institute. His current research focuses on the merging between AI and communication networks. Dr. Cui is a Fellow of the Canadian Academy of Engineering and the Royal Society of Canada. He is a Member of the Steering Committee for IEEE Transactions on Big Data. He is the Editor-in-Chief for IEEE Transactions on Mobile Computing, an Area Editor for IEEE Signal Processing Magazine, and an Associate Editor for IEEE Transactions on Big Data, IEEE Transactions on Signal Processing, IEEE Journal on Selected Areas in Communications, IEEE Transactions on Green Communications and Networking, and IEEE Transactions on Wireless Communications. He is the Chair of the Steering Committee for IEEE Transactions on Cognitive Communications and Networking. He is also the Vice Chair of the IEEE VT Fellow Evaluation Committee and a Member of the IEEE ComSoc Award Committee. He was an Elected Member of the IEEE Signal Processing Society SPCOM Technical Committee (2009–2014) and the Elected Chair of the IEEE ComSoc Wireless Technical Committee (2017–2018). He was selected as a Thomson Reuters Highly Cited Researcher and listed in the World’s Most Influential Scientific Minds by ScienceWatch in 2014. He was elected as an IEEE ComSoc Distinguished Lecturer in 2014 and the IEEE VT Society Distinguished Lecturer in 2019
\end{IEEEbiography}

\begin{IEEEbiography}[{\includegraphics[width=1in,height=1.25in,clip,keepaspectratio]{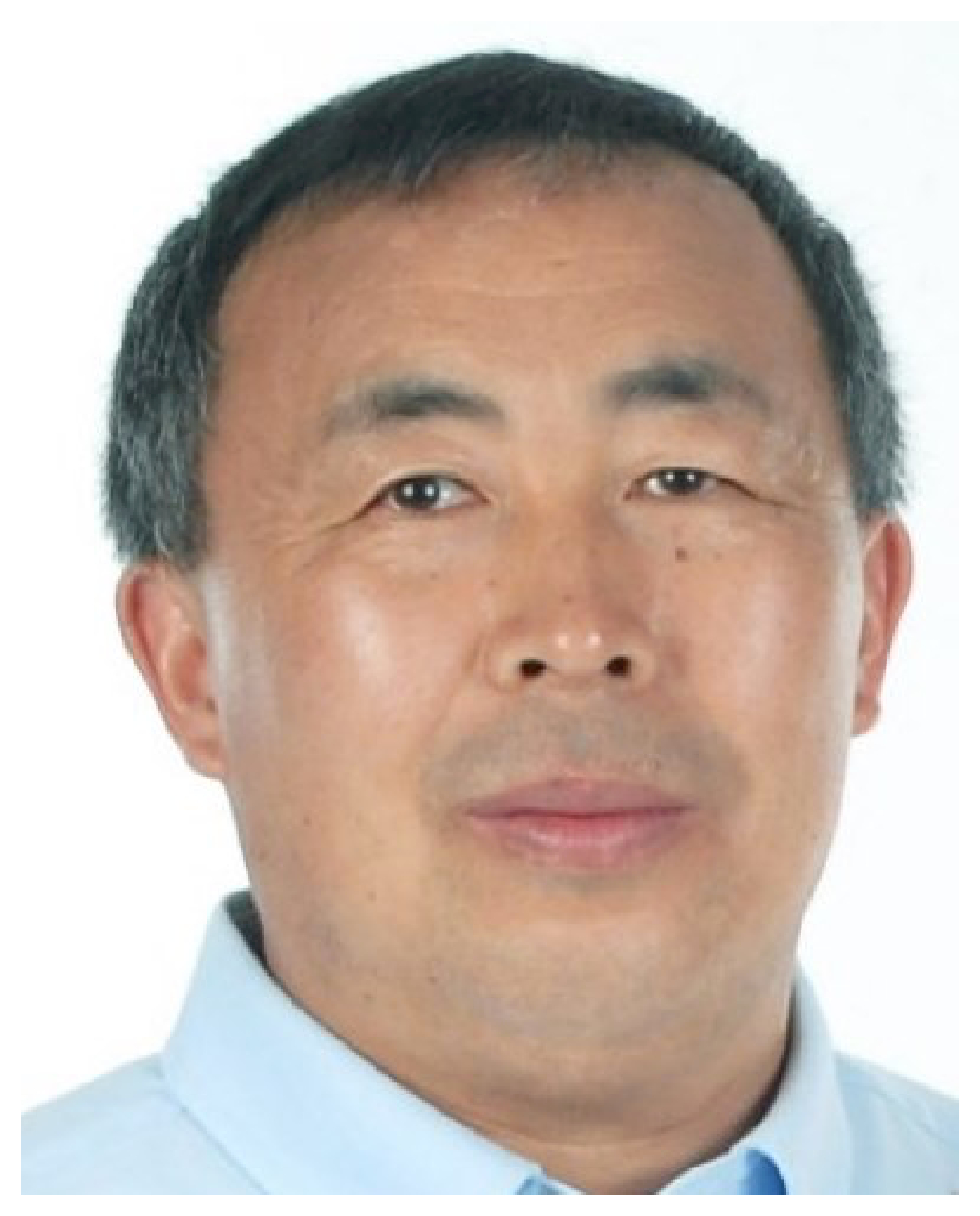}}]{Xuemin (Sherman) Shen}
received the Ph.D. degree in electrical engineering from Rutgers University, New Brunswick, NJ, USA, in 1990. He is a University Professor with the Department of Electrical and Computer Engineering, University of Waterloo, Canada. His research focuses on network resource management, wireless network security, Internet of Things, 5G and beyond, and vehicular networks. Dr. Shen is a registered Professional Engineer of Ontario, Canada, an Engineering Institute of Canada Fellow, a Canadian Academy of Engineering Fellow, a Royal Society of Canada Fellow, a Chinese Academy of Engineering Foreign Member, and a Distinguished Lecturer of the IEEE Vehicular Technology Society and Communications Society. 
 
Dr. Shen received “West Lake Friendship Award” from Zhejiang Province in 2023, President's Excellence in Research from the University of Waterloo in 2022,  the Canadian Award for Telecommunications Research from the Canadian Society of Information Theory (CSIT) in 2021, the R.A. Fessenden Award in 2019 from IEEE, Canada, Award of Merit from the Federation of Chinese Canadian Professionals (Ontario) in 2019, James Evans Avant Garde Award in 2018 from the IEEE Vehicular Technology Society, Joseph LoCicero Award in 2015 and Education Award in 2017 from the IEEE Communications Society (ComSoc), and Technical Recognition Award from Wireless Communications Technical Committee (2019) and AHSN Technical Committee (2013). He has also received the Excellent Graduate Supervision Award in 2006 from the University of Waterloo and the Premier’s Research Excellence Award (PREA) in 2003 from the Province of Ontario, Canada. He serves/served as the General Chair for the 6G Global Conference’23, and ACM Mobihoc'15, Technical Program Committee Chair/Co-Chair for IEEE Globecom’24, 16 and 07, IEEE Infocom’14, IEEE VTC’10 Fall, and the Chair for the IEEE ComSoc Technical Committee on Wireless Communications. Dr. Shen is the President of the IEEE ComSoc. He was the Vice President for Technical \& Educational Activities, Vice President for Publications, Member-at-Large on the Board of Governors, Chair of the Distinguished Lecturer Selection Committee, and Member of the IEEE Fellow Selection Committee of the ComSoc. Dr. Shen served as the Editor-in-Chief of the IEEE IoT JOURNAL, IEEE Network, and IET Communications.
\end{IEEEbiography}

\ifCLASSOPTIONcaptionsoff
  \newpage
\fi

\end{document}